\RequirePackage{fix-cm}
\documentclass[10pt,journal,compsoc]{IEEEtran}

\usepackage{amsmath, amsfonts, amssymb}
\usepackage{array}
\usepackage{textcomp}
\usepackage{stfloats}

\usepackage{graphicx}
\usepackage[caption=false, font=normalsize, labelfont=sf, textfont=sf]{subfig}

\usepackage{booktabs}
\usepackage{multirow}
\usepackage{tabularx}
\usepackage{nicematrix}
\usepackage{threeparttable}
\usepackage{pifont}

\usepackage{algorithm}
\usepackage{algorithmic}

\usepackage{enumitem}
\usepackage{verbatim}
\usepackage{url}

\usepackage{cite}

\usepackage{xcolor}
\usepackage{fontawesome5}

\definecolor{starcolor}{RGB}{255, 193, 7}

\newcommand{\starY}[1]{
    \textcolor{starcolor}{
        \ifcase#1\or \faStar\or \faStar\faStar\or \faStar\faStar\faStar\fi
    }
}

\newcommand{\subsubsubsection}[1]{\vspace{0.1em}\begin{itemize}[leftmargin=*]\item\textbf{#1}\end{itemize}}

\newcolumntype{C}{>{\centering\arraybackslash}X}
\newcolumntype{P}[1]{>{\centering\arraybackslash}p{#1}}

\usepackage{hyperref}

\usepackage{eso-pic}

\newcommand{\IEEECopyrightNotice}{
\fontsize{6.5}{7}\selectfont
\parbox{0.9\paperwidth}{
\centering
© 2026 IEEE. Personal use of this material is permitted. Permission from IEEE must be obtained for all other uses, in any current or future media, including reprinting/republishing this material for advertising or promotional purposes, creating new collective works, for resale or redistribution to servers or lists, or reuse of any copyrighted component of this work in other works.
}
}

\AddToShipoutPictureFG*{
  \AtPageLowerLeft{
    \ifnum\value{page}=1
      \raisebox{0.8cm}[0pt][0pt]{
        \makebox[\paperwidth]{\IEEECopyrightNotice}
      }
    \fi
  }
}

\begin{document}

\title{A Survey on Interpretability in Visual Recognition}

\author{Qiyang Wan,~\IEEEmembership{Student Member,~IEEE,}
Chengzhi Gao,~\IEEEmembership{Student Member,~IEEE,}

Ruiping Wang,~\IEEEmembership{Senior Member,~IEEE,}
Xilin Chen,~\IEEEmembership{Fellow,~IEEE}
\thanks{Q. Wan, C. Gao, R. Wang, and X. Chen are with the Key Laboratory of AI Safety of Chinese Academy of Sciences (CAS), Institute of Computing Technology, CAS, Beijing 100190, China. Email: \{qiyang.wan, chengzhi.gao\}@vipl.ict.ac.cn, \{wangruiping, xlchen\}@ict.ac.cn}
}

\bstctlcite{IEEEtranControl}

\IEEEtitleabstractindextext{
\begin{abstract}
Visual recognition models have achieved unprecedented success in various tasks.
While researchers aim to understand the underlying mechanisms of these models, the growing demand for deployment in safety-critical areas like autonomous driving and medical diagnostics has accelerated the development of eXplainable AI (XAI).
Distinct from generic XAI, visual recognition XAI is positioned at the intersection of vision and language, which represent the two most fundamental human modalities and form the cornerstones of multimodal intelligence.
This paper provides a systematic survey of XAI in visual recognition by establishing a multi-dimensional taxonomy from a human-centered perspective based on \textbf{intent}, \textbf{object}, \textbf{presentation}, and \textbf{methodology}.
Beyond categorization, we summarize critical evaluation desiderata and metrics, conducting an extensive qualitative assessment across different categories and demonstrating quantitative benchmarks within specific dimensions.
Furthermore, we explore the interpretability of Multimodal Large Language Models and practical applications, identifying emerging trends and opportunities.
By synthesizing these diverse perspectives, this survey provides an insightful roadmap to inspire future research on the interpretability of visual recognition models.
\end{abstract}

}

\maketitle

\section{Introduction}
\label{sec:intro}

\IEEEPARstart{m}{ethods} for visual recognition have undergone extensive development and have been successfully applied across various domains.
Furthermore, researchers are increasingly investigating the underlying mechanisms driving these systems, an area referred to as interpretability research.
This paper presents a systematic review of interpretable visual recognition.
We provide a holistic framework designed to enable researchers and developers, including those without prior knowledge of interpretability, to intuitively understand the characteristics of various interpretable visual recognition approaches.

\subsection{Background}
\label{sec:intro-background}

The rapid deployment of visual recognition models has revolutionized fields such as healthcare diagnostics, autonomous vehicles, and surveillance systems.
However, the inherent opaque nature of deep learning models often obscures the relationship between inputs and specific outputs.
As these models assume critical roles in human society, the lack of transparency in their decision making processes poses significant ethical and safety risks.
This challenge has catalyzed the field of e\textbf{X}plainable \textbf{A}rtificial \textbf{I}ntelligence, which strives to elucidate the behaviors and decision boundaries of complex algorithms.
Beyond diagnosing model failures, interpretability is essential for enhancing human trust and facilitating effective human computer interaction~\cite{ferguson2021reframing}.

\begin{figure}[t]
\centering
\includegraphics[width=0.9\linewidth]{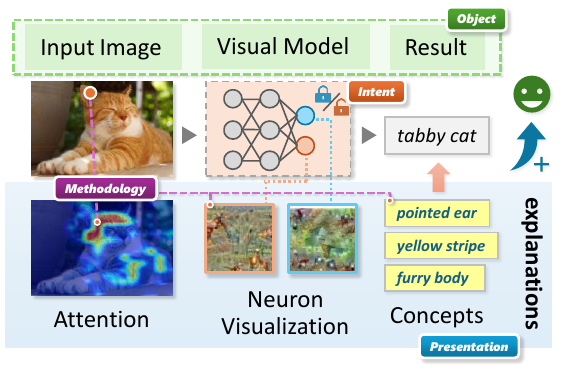}
\caption{Overview of XAI in visual recognition.
While black-box models provide only raw predictions, interpretability research offers diverse explanations to enhance human trust.
The taxonomy proposed in this survey groups existing methods along four dimensions: \textbf{intent}, \textbf{object}, \textbf{presentation}, and \textbf{methodology}.}
\label{fig:xai}
\end{figure}

Visual recognition is the most fundamental task in a multimodal systems, with its accuracy and robustness being critical to the performance of subsequent higher-level tasks.
It is uniquely situated at the intersection of visual perception and semantic concepts.
As illustrated in Fig.~\ref{fig:xai}, visual recognition follows a standardized pipeline that transforms visual signals into semantic category labels.
The variability in inputs and outputs substantially increases the complexity of XAI research in the domain of visual recognition.

As depicted in Fig.~\ref{fig:xai}, techniques such as activation heatmap, neuron visualization, and concept bottleneck have been proposed to offer understandable explanations behind predictions.
However, whether to provide explanations and what kind of explanations to provide can have either positive or negative effects on human trust~\cite{papenmeier2022s,kim2023help}.
To establish a structured understanding of this evolving field and address its inherent complexity, this survey proposes a taxonomy organized along four human-centered dimensions:
\begin{itemize}[leftmargin=*]
\item \textbf{Intent:} \textit{What is the purpose of bringing in interpretability?}
\item \textbf{Object:} \textit{What does the generated explanation focus on?}
\item \textbf{Presentation:} \textit{What does the generated explanation look like?}
\item \textbf{Methodology:} \textit{How is the explanation generated?}
\end{itemize}

Beyond the core taxonomy, this survey further explores additional critical facets of the field as outlined in Fig. 2.
We provide a thorough review of evaluation desiderata and metrics to address the comparison of different approaches.
The survey then investigates XAI in multimodal large language models to identify emerging tools and interpretability techniques for complex models.
Finally, we summarize practical applications and discuss future research opportunities to provide an insightful roadmap for the community.
By synthesizing these diverse components, this paper offers a timely framework for understanding and advancing the interpretability of visual recognition.

\begin{figure*}[t]
\centering
\includegraphics[width=\linewidth]{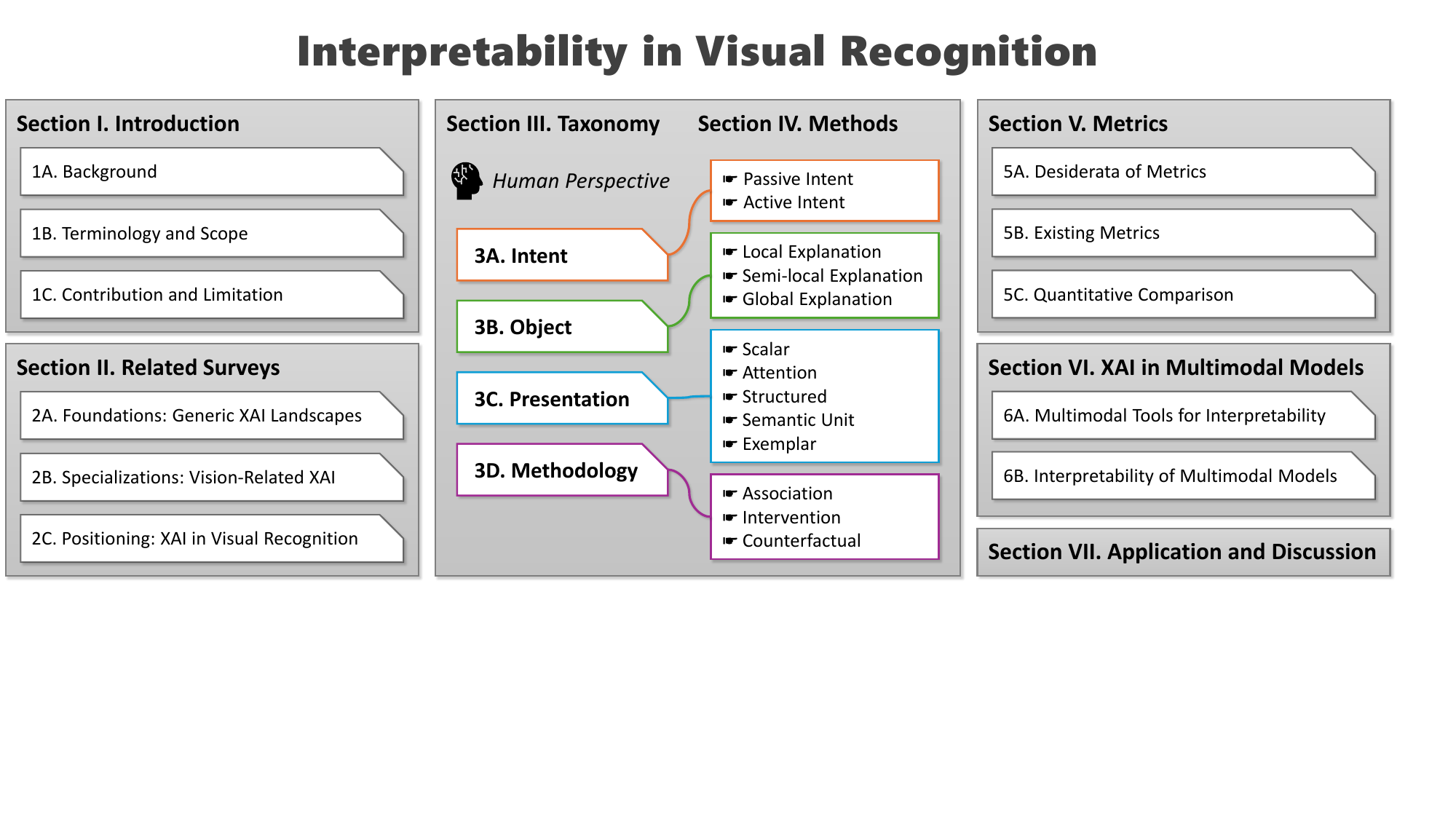}
\caption{Structural organization of the survey.
The primary contribution of this paper lies in proposing a multi-dimensional taxonomy to describe the interpretability of visual recognition across \textbf{intent}, \textbf{object}, \textbf{presentation}, and \textbf{methodology}.
As detailed in Sec.~\ref{sec:taxonomy}, this framework categorizes existing methods from a human centered perspective to enhance systematic understanding.
Beyond this taxonomy, the survey encompasses evaluation metrics, multimodal large language models, and practical applications to provide an insightful research roadmap.
}
\label{fig:structure}
\end{figure*}

\subsection{Terminology and Scope}
\label{sec:intro-definition}

XAI research encompasses processes and methods aimed at making AI models’ internal logic and outputs human-understandable.
The primary motivation arises from the ``black-box'' nature of models lacking inherent interpretability, where complex structures obscure their decision logic~\cite{rudin2022interpretable}.
XAI research typically follows two paradigms: post-hoc analysis, which interrogates a trained black-box model using techniques like visualization or perturbation; and intrinsic design, which integrates transparent modules directly into the model's architecture.
While some studies~\cite{leblanc2023relationship} distinguish between \textbf{explainability} (post-hoc) and \textbf{interpretability} (intrinsic), most literature uses these terms interchangeably, so this survey treats them as equivalent.
To avoid ambiguity, we utilize the terms \textbf{post-hoc} and \textbf{self-interpretable} when emphasizing their differences.

This paper primarily investigates visual recognition models, which map an input image $\mathbf{x}$ to a prediction $\hat{y}$ through a backbone feature extractor $f$ and a classifier head $g$, such that $\hat{y} = g(f(\mathbf{x}))$.
Current research predominantly focuses on the classifier $g$ and image features $\mathbf{z}=f(\mathbf{x})$, whereas studies on the backbone $f$ remain in the early stages and primarily target the top layers, as these layers are more likely to possess semantic information.
For both post-hoc and self-interpretable paradigms, the explanations provided to researchers, developers, or users are typically presented external to the recognition pipeline and are highly diverse.
Due to the inherent coupling among visual tasks, our discussion naturally extends to localization-based interpretability in detection and segmentation, as well as semantic interactions within multimodal frameworks.
Specifically, as XAI in visual recognition bridges both visual and textual modalities, its interpretability frequently encompass both \textbf{localization interpretability} and \textbf{semantic interpretability}.

\subsection{Contribution and Limitation}
\label{sec:intro-contrib}

This survey distinguishes itself from prior literature in two primary aspects: it focuses exclusively on XAI research tailored for visual recognition models, and it systematically organizes relevant XAI methods from a multi-dimensional, human-centered perspective.
Given that XAI is an expansive field, surveys with an overly broad scope often suffer from a lack of granular focus and practical applicability.
By narrowing the scope to visual recognition, this paper provides a more detailed, task-oriented, and practical classification of the methods.
The proposed multi-dimensional framework enables practitioners to efficiently grasp advancements and locate suitable methods for specific applications.
However, the the taxonomy is specifically tailored to visual recognition models, and expanding this taxonomy to encompass a broader spectrum of vision-related tasks presents inherent challenges.
Moreover, while current design was designed with extensibility in mind, its current focus remains primarily on visual and textual modalities; consequently, the framework may require significant adaptation as novel explanatory modalities emerge.
Addressing these limitations remains a subject for future work, requiring further research to effectively adapt and scale the proposed framework to more complex scenarios.

\begin{table*}[t]
\caption{Summary of recent related XAI surveys}
\label{tab:rs-a}
\begin{NiceTabular}{c>{\hspace{-2pt}}c<{\hspace{-2pt}}ccX}[width=\textwidth,cell-space-limits=0.5ex]
\CodeBefore
\rowcolor{gray!20}{1}
\rowlistcolors{2}{white,gray!10}[restart,cols={2-5}]
\Body
\toprule
\textbf{Topic} & \textbf{Ref.} & \textbf{Year} & \textbf{Literature} & \textbf{Description} \\
\midrule
\Block[v-center]{11-1}{Generic \\[0.2em] XAI} & \cite{kabir2025review} & 2025 & 2010-2025 & Survey XAI methodologies, applications, and open challenges toward trustworthy AI \\
& \cite{schwalbe2024comprehensive} & 2024 & 2017-2024 & Propose a unified taxonomy of XAI methods and provide use-case-oriented insights \\
& \cite{hassija2024interpreting} & 2024 & 2016-2024 & Review XAI models, evaluation metrics, challenges, and trends to enhance transparency \\
& \cite{saeed2023explainable} & 2023 & 2017-2022 & Conduct a systematic meta-survey of XAI challenges and future research \\
& \cite{dwivedi2023explainable} & 2023 & 2017-2021 & Survey XAI techniques and guide framework selection for interpretable AI systems \\
& \cite{ras2022explainable} & 2022 & 2014-2022 & Discuss key methods, evaluations, and future directions in explainable deep learning \\
& \cite{rudin2022interpretable} & 2022 & 2013-2022 & Identify 10 technical challenge areas and provide historical and background context \\
& \cite{zhang2021survey} & 2021 & 2015-2021 & Propose a taxonomy of neural network interpretability based on engagement, type, and focus \\
& \cite{samek2021explaining} & 2021 & 2014-2021 & Review post-hoc explanations, evaluate XAI methods, and demonstrate applications \\
& \cite{das2020opportunities} & 2020 & 2017-2020 & Review XAI techniques, including taxonomy, methods, principles, and evaluation \\
& \cite{arrieta2020explainable} & 2020 & 2007-2020 & Explore the importance of explainability in AI and present a taxonomy of XAI techniques \\
\midrule
\Block[v-center]{5-1}{Multimodal \\[0.2em] XAI} & \cite{kazmierczak2025explainability} & 2025 & 2017-2025 & Survey integration of foundation models with explainable AI in the vision domain \\
& \cite{saleh2025building} & 2025 & 2013-2025 & Analyze fairness, explainability, and ethics in vision–language models across multimodal tasks \\
& \cite{sun2024review} & 2024 & 1999-2024 & Systematize MXAI across model eras and summarize metrics, datasets, and open challenges \\
& \cite{dang2024explainable} & 2024 & 2017-2024 & Analyze recent advances in Multimodal XAI, focusing on methods, datasets, and metrics \\
& \cite{rodis2024multimodal} & 2024 & 2016-2024 & Survey interpretability of MLLMs, categorizing evaluations and future directions \\
\midrule
\Block[v-center]{4-1}{XAI in \\[0.2em] Visual Task} & \cite{cheng2025comprehensive} & 2025 & 2008-2025 & Provide a comparative evaluation and hierarchical taxonomy of visual XAI methods \\
& \cite{gipivskis2024explainable} & 2024 & 2017-2024 & Survey XAI in semantic segmentation, categorizing evaluation metrics and future challenges \\
& \cite{bai2021explainable} & 2021 & 2015-2021 & Review explainable deep learning, efficiency, and robustness in pattern recognition \\
& \cite{he2021interpretable} & 2021 & 2017-2021 & Systematize visual reasoning by explanation forms, datasets, challenges, and future directions \\
\midrule
\Block[v-center]{4-1}{XAI by \\[0.2em] Visualization} & \cite{bhati2025survey} & 2025 & 2015-2025 & Systematize post-hoc XAI methods with a taxonomy spanning local and concept explanations \\
& \cite{baniecki2024adversarial} & 2024 & 2017-2024 & Survey adversarial attacks on XAI, outlining security challenges and suggesting directions \\
& \cite{abhishek2022attribution} & 2022 & 2011-2022 & Review attribution XAI techniques and identify challenges in robust explanation and evaluation \\
& \cite{alicioglu2022survey} & 2022 & 2018-2021 & Review trends and challenges in visual analytics for XAI \\
\midrule
\Block[v-center]{3-1}{XAI about \\[0.2em] Architecture} & \cite{fantozzi2024explainability} & 2024 & 2017-2024 & Survey transformer explainability, categorizing by components, applications, and visualization \\
& \cite{kashefi2023explainability} & 2023 & 2021-2023 & Review XAI methods for vision transformers, categorizing approaches and evaluation criteria \\
& \cite{ibrahim2023explainable} & 2023 & 2011-2023 & Systematize explainable CNN approaches and review metrics, applications, and open challenges \\
\bottomrule
\end{NiceTabular}
\end{table*}

\section{Related Surveys}
\label{sec:rw}

Many surveys have focused on organizing the literature related to XAI.
In this section, we categorize these surveys into two primary groups based on their relevance to our subject: generic XAI and specific vision-related XAI.
Some of the surveys \footnote{Due to space limitations, the complete tables are available at \\ \url{https://vipl-vsu.github.io/xai-recognition/}.} are summarized in Tab.~\ref{tab:rs-a}.

Given the explosion of interest in interpretability, a significant number of surveys address XAI from a broad perspective.
We select influential works from each year to reflect the evolution of mainstream thought, as illustrated in the first part of Tab.~\ref{tab:rs-a}.
Early foundational surveys~\cite{arrieta2020explainable, das2020opportunities, samek2021explaining, zhang2021survey} focused on defining fundamental taxonomies (e.g., post-hoc vs. intrinsic), while recent works address more key challenges~\cite{kabir2025review, hassija2024interpreting, saeed2023explainable, rudin2022interpretable}, evaluation frameworks~\cite{hassija2024interpreting, ras2022explainable}, systems~\cite{dwivedi2023explainable}, and future directions~\cite{kabir2025review, hassija2024interpreting, saeed2023explainable, ras2022explainable} of XAI.

From these surveys, several consensus conclusions regarding current challenges can be drawn, such as the trade-off between model performance and transparency, a lack of unified evaluation metrics, etc.
While these generic surveys offer high-level conceptual frameworks and guidelines that apply across machine learning, their broad scope often results in a lack of practical depth for specific domains.
They provide the ``what'' and ``why'' of XAI but frequently fall short on the ``how''when applied to complex, modality-specific data and tasks.

In contrast to broad, general-purpose surveys, research on XAI in the vision domain offers more fine-grained and domain-specific insights by systematically considering the distinctive spatial and structural characteristics of visual data.
Accordingly, we categorize these specialized surveys into several sub-domains:

\begin{itemize}[leftmargin=*]
\item \textbf{Multimodal XAI:} These recent works address the intersection of vision and language, exploring how Multimodal Large Language Models (MLLMs) justify cross-modal reasoning~\cite{kazmierczak2025explainability, rodis2024multimodal} and corresponding evaluation frameworks~\cite{dang2024explainable, sun2024review}.
\item \textbf{XAI in Visual Task:} These works investigate XAI within specific visual tasks, such as pattern recognition~\cite{bai2021explainable}, semantic segmentation~\cite{gipivskis2024explainable}, and visual reasoning~\cite{he2021interpretable}, focusing on how explanations can improve task-specific trust.
\item \textbf{XAI by Visualization:} These surveys concentrate on the technical mechanisms of ``seeing'' into the black box, covering visual analytics~\cite{alicioglu2022survey}, attribution-based methods~\cite{bhati2025survey, abhishek2022attribution}, and the security of visual explanations~\cite{baniecki2024adversarial}.
\item \textbf{XAI about Architecture:} With the rise of complex backbones, these surveys analyze interpretability through the lens of specific structures, such as Convolutional Neural Networks (CNNs)~\cite{ibrahim2023explainable} or the emerging explainability of Vision Transformers (ViTs)~\cite{kashefi2023explainability, fantozzi2024explainability}.
\end{itemize}

These specialized surveys are significantly more detailed than generic ones, offering practical value for researchers dealing with specific architectures or data modalities.
Our work belongs to this category but identifies a critical gap: \textbf{Visual Recognition}. 
As the most fundamental cognitive task in computer vision, robust recognition serves as the bedrock for higher-level tasks like detection, segmentation, or reasoning.
Unlike existing surveys that may focus strictly on a single technique (like visualization) or a single architecture (like ViTs), this survey provides an overview of XAI methods specifically through the lens of visual recognition, including ones with various backbones and methodologies.
To the best of our knowledge, this is the first survey to aggregate and analyze XAI methods at the specific granularity of visual recognition, bridging the gap between low-level technical methods and high-level task requirements.

\section{Taxonomy}
\label{sec:taxonomy}

\begin{figure}[t]
\centering
\includegraphics[width=\linewidth]{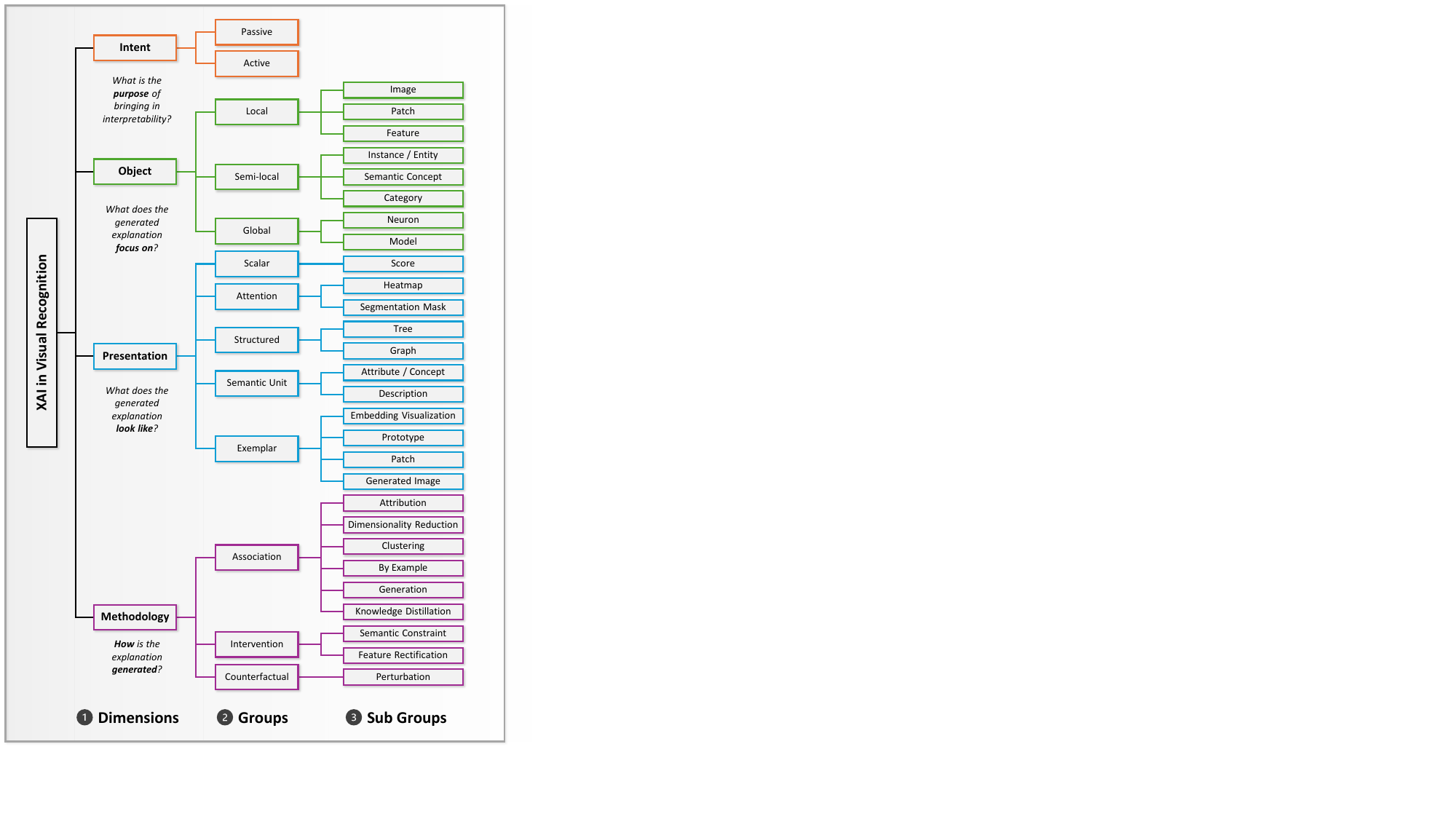}
\caption{The proposed taxonomy and corresponding method groups of XAI in visual recognition.}
\label{fig:framework}
\end{figure}

As discussed in Sec.~\ref{sec:rw}, XAI for \textbf{visual recognition} represents a pragmatic level of abstraction that preserves a holistic view while enabling the analysis of fine-grained technical details.
Accordingly, we propose a taxonomy that provides a practical and systematic framework for organizing existing methods in this area.
Building on community consensus and the distinctive demands of visual recognition (Fig.~\ref{fig:xai}), we identify \textbf{intent}, \textbf{object}, \textbf{presentation}, and \textbf{methodology} as the four most critical dimensions, and leverage them to reorganize interpretable methods into a human-intuitive framework, as illustrated in Fig.~\ref{fig:framework}.
This taxonomy defines clear semantics and classification rules for each dimension, enabling a natural grouping of methods that serves as an effective index for diverse interpretability requirements.
We then introduce each dimension in turn, and discuss representative methods in Sec.~\ref{sec:method} for better understanding.

The \textbf{intent} of interpretability describes the purpose for introducing interpretability into visual recognition models.
It is widely adopted in the XAI literature and serves as a fundamental organizing principle.
It is commonly categorized into two types: \textbf{passive} (or post-hoc) and \textbf{active} (or intrinsic, self-interpretable).
\textbf{Passive interpretability} encompasses methods that do not alter the recognition model itself, but instead explain an already trained model by analyzing its internal mechanisms or decision process.
In contrast, \textbf{active interpretability} incorporates interpretability into the model design, making the recognition process inherently transparent.
% This dichotomy is widely adopted in the XAI literature and serves as a fundamental organizing principle.

The \textbf{object} of interpretability refers to the target within the recognition pipeline that an explanation aims to address, which can be viewed as the input to the explanation module (green blocks in Fig.~\ref{fig:xai}).
Different application scenarios impose different explanatory needs.
For example, in medical image analysis, clinicians typically require explanations for individual diagnostic outcomes, resulting in \textbf{local explanations} at the sample level.
In contrast, tasks that focus on shared visual characteristics across a class of samples, corresponding to \textbf{semi-local explanations}.
Finally, in high-stakes or safety-critical settings, it may be necessary to explain the complete decision logic of the model in a category-independent manner, yielding \textbf{global explanations}.
It is also broadly recognized across XAI research, including but not limited to visual recognition~\cite{ras2022explainable,hassija2024interpreting,dwivedi2023explainable,saranya2023systematic,schwalbe2024comprehensive,hohl2024opening,borys2023explainable,weber2023beyond,nazir2023survey}.

The \textbf{presentation} of interpretability characterizes how explanations are conveyed to users, corresponding to the outputs of XAI methods (blue blocks in Fig.~\ref{fig:xai}).
This dimension highlights a key distinction between visual XAI and generic XAI.
Since visual recognition inherently bridges visual inputs and semantic labels, interpretability research naturally spans both visual and textual modalities.
Visual explanations often decompose images into spatially localized regions, whereas textual explanations break down category predictions into semantic concepts.
These perspectives give rise to two dominant paradigms: \textbf{localization interpretability} and \textbf{semantic interpretability}.
Specifically, explanation presentations can take diverse forms, including \textbf{scalar}, \textbf{attention}, \textbf{structured representation}, \textbf{semantic unit}, \textbf{exemplar}, and related variants.

The \textbf{methodology} of interpretability describes how explanations are derived.
Based on their causal assumptions and influence on the model, methods can be aligned with \textit{The Ladder of Causation}~\cite{pearl2018book} and grouped into \textbf{association}, \textbf{intervention}, and \textbf{counterfactual}.
Association-based methods capture correlations and are commonly used for passive interpretability; intervention-based methods analyze outcomes under active modifications, supporting active interpretability; counterfactual methods simulate alternative inputs and are particularly suited to black-box models.
Importantly, the \textbf{methodology} dimension is closely intertwined with the other dimensions and should not be considered in isolation.
Once the \textbf{intent}, \textbf{object}, and \textbf{presentation} are specified, the choice of feasible interpretability technologies is largely determined.

\begin{table}[t]
\centering
\begin{threeparttable}
\caption{Multi-dimensional taxonomies proposed by the previous surveys}
\label{tab:taxonomy-comparison}
\setlength{\tabcolsep}{3pt}
\renewcommand{\arraystretch}{1.1}
\begin{tabularx}{\linewidth}{lCCCC}
\toprule
Survey Topic & \textit{Intent} & \textit{Object} & \textit{Present.} & \textit{Method.} \\
\midrule
\multicolumn{5}{l}{\textbf{Broader Scope}} \\
\quad Generic XAI~\cite{schwalbe2024comprehensive} & \ding{51} & \ding{51} & \ding{70} & \ding{70} \\
\quad Multimodal XAI~\cite{rodis2024multimodal} & \ding{51} & \ding{51} & \ding{70} & \ding{70} \\
\addlinespace[0.1em]
\multicolumn{5}{l}{\textbf{Narrower Scope}} \\
\quad Attribution (Technology)~\cite{cheng2025comprehensive} & \ding{55} & \ding{55} & \ding{70} & \ding{51} \\
\quad CNN (Architecture)~\cite{ibrahim2023explainable} & \ding{51} & \ding{51} & \ding{55} & \ding{70} \\
\midrule
\textbf{Visual Recognition (Ours)} & \ding{51} & \ding{51} & \ding{51} & \ding{51} \\
\bottomrule
\end{tabularx}
\begin{tablenotes}
    \footnotesize
    \item \ding{70} indicates that the dimension is partially discussed.
\end{tablenotes}
\end{threeparttable}
\end{table}

With the proposed four-dimensional taxonomy, we aim to provide a structured and task-aware view of interpretability in visual recognition.
To better position this taxonomy within the broader XAI literature, we briefly review representative taxonomic frameworks proposed in previous surveys and summarize their coverage in Tab.~\ref{tab:taxonomy-comparison}.
Surveys with a broader scope, such as generic XAI~\cite{schwalbe2024comprehensive} and multimodal XAI~\cite{rodis2024multimodal}, typically emphasize high-level conceptual distinctions.
In particular, widely adopted dichotomies such as post-hoc versus intrinsic explanations (\textbf{intent}) and global versus local explanations (\textbf{object}) are consistently highlighted, while \textbf{presentation} and \textbf{methodology} are discussed at a more abstract level.
In contrast, surveys with a narrower scope concentrate on specific technical aspects, such as attribution-based explanations~\cite{cheng2025comprehensive} or CNN-centric interpretability~\cite{ibrahim2023explainable}.
These works provide fine-grained analyses of selected dimensions, yet naturally limit their coverage to a subset of the overall interpretability landscape.

Overall, Tab.~\ref{tab:taxonomy-comparison} illustrates that existing taxonomies capture complementary perspectives at different levels of abstraction.
The proposed four-dimensional taxonomy offers a unified view of interpretability in visual recognition, and provides a coherent organizational structure for the method-level review in the following sections.

\section{Methods}
\label{sec:method}

In this section, we will introduce groups and specific values for each dimension mentioned above and provide the representative methods.
According to Sec.~\ref{sec:taxonomy}, we introduce the methods from \textbf{intent}, \textbf{object}, \textbf{presentation}, and \textbf{methodology} respectively.
It's important to note that the proposed taxonomy is not a tree structure, but tags each work on various dimensions.
Additionally, even within a single dimension, the values are mostly non-exclusive.
Therefore, a method may appear in different sections, and we will discuss one method from multiple perspectives to help readers better understand the proposed taxonomy.

\subsection{Intent}
\label{sec:method-intent}

From the perspective of \textbf{intent}, interpretability in visual recognition can be broadly categorized into \textbf{passive} and \textbf{active} approaches, depending on whether interpretability is applied post hoc or explicitly incorporated into the model design.

\subsubsection{Passive}
\label{sec:method-intent-passive}   

\textbf{Passive} interpretability focuses on explaining a trained model’s predictions without altering its architecture, making it particularly suitable for analyzing complex black-box systems.
Representative techniques include attribution-based methods such as CAM~\cite{zhou2016learning}, Grad-CAM~\cite{selvaraju2017grad}, Integrated Gradients~\cite{sundararajan2017axiomatic}, LRP~\cite{bach2015pixel}, and SmoothGrad~\cite{smilkov2017smoothgrad}, which generate saliency maps to highlight input regions that contribute most to a decision.
Perturbation-based approaches, such as explaining prototypes~\cite{nauta2021looks} and counterfactual reasoning methods like CaCE~\cite{goyal2019explaining}, instead analyze model behavior through systematic input modifications.
While passive methods are model-agnostic and easy to deploy, their explanations are derived independently from the prediction process and may not faithfully reflect the model’s true reasoning, often resulting in fragile or misleading interpretations~\cite{rudin2022interpretable}.

\subsubsection{Active}    
\label{sec:method-intent-active}  

\textbf{Active} interpretability embeds interpretability directly into the model architecture or training procedure, ensuring that predictions and explanations are generated simultaneously.
Typical examples include Concept Bottleneck Models~\cite{koh2020concept}, which enforce decision-making through human-interpretable concepts, and prototype-based frameworks such as ProtoPNet~\cite{chen2019looks}, ProtoTree~\cite{nauta2021neural}, and ProtoPool~\cite{rymarczyk2022interpretable}, where predictions rely on learned, interpretable prototypes.
Other approaches intervene at the training level, for instance Interpretable CNNs~\cite{zhang2018interpretable,zhang2020interpretable}, which introduce specialized losses to encourage semantic alignment in convolutional layers.
By aligning interpretability with the model’s internal mechanisms, active methods enhance transparency and user trust, especially in high-stakes scenarios~\cite{rudin2022interpretable}; however, they may restrict model expressiveness and incur a trade-off between interpretability and predictive performance.

\subsection{Object}
\label{sec:method-object}

\begin{figure}[t]
\centering
\includegraphics[width=\linewidth]{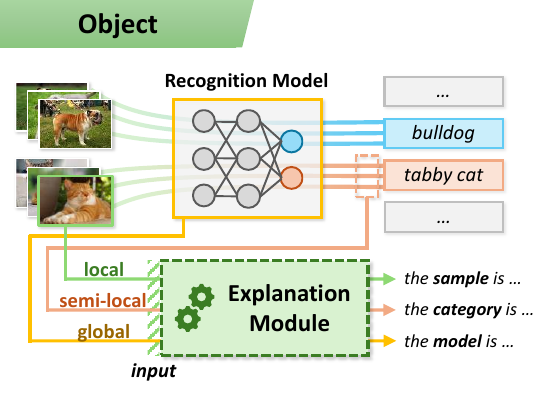}
\caption{Illustration of \textbf{Object}.
XAI methods can be categorized as \textbf{local} or \textbf{global}, depending on whether the explanation module receives a single sample or the entire model as input.
Specifically, in the context of visual recognition, it is also important to consider the model’s representations of categories, concepts, and other high-level semantic labels, which may be viewed as \textbf{semi-local} explanations.}
\label{fig:object}
\end{figure}

As illustrated in Fig.~\ref{fig:object}, according to the \textbf{object} of the visual XAI methods, they can be categorized into \textbf{local} and \textbf{global}, which respectively refer to explanations for a single sample and the entire model.
Additionally, in visual recognition, the embedded high-level semantic labels within models, such as categories and concepts that represent specific groups of samples, have received considerable attention.
Explanations that describe these semantic labels are referred to as \textbf{semi-local} explanations.

\subsubsection{Local}

\textbf{Local} interpretability refers to explanations that focus on individual samples.
Sample-level inputs do not refer solely to input images; intermediate results such as patches and features are also obviously sample-level.
Local explanations can mainly be further categorized into the following types:

\subsubsubsection{Image}   

Image are the most common object for local explanation methods~\cite{selvaraju2017grad,zhou2016learning,sundararajan2017axiomatic,bach2015pixel,smilkov2017smoothgrad,montavon2017explaining,park2018multimodal,zhou2018interpretable,ghandeharioun2021dissect,wang2023learning2,achtibat2023attribution,fel2023don,schrouff2021best,sun2024explain,li2021visualizing,dreyer2023understanding,qiang2023interpretability}.
These methods usually generate an explanation for a single image, which serve as the primary input to visual recognition models.
Most attribution methods take image as the object of interpretation, aiming to represent the distribution of a model's attention on an input image.
Methods, such as CAM \cite{zhou2016learning}, GradCAM \cite{selvaraju2017grad}, IG \cite{sundararajan2017axiomatic}, LRP \cite{bach2015pixel}, SmoothGrad \cite{smilkov2017smoothgrad}, etc., generate visual explanations that highlight the regions of an image most influential in the model's decision.
These interpretations are tailored to the particular input image, making them highly specific and useful for understanding the model’s decision for that image alone.

\subsubsubsection{Patch}

People are usually interested in how specific image patches influence the model's decisions~\cite{chen2019looks,nauta2021neural,hase2019interpretable,hernandez2021natural,nauta2021looks,nauta2023pip,wang2023learning,li2021visualizing,ma2024looks,hagos2023distance}.
The most representative methods are models relying on prototypical parts such as ProtoPNet \cite{chen2019looks}, ProtoTree \cite{nauta2021neural}, PIP-Net \cite{nauta2023pip}, ProtoConcepts \cite{ma2024looks} etc.
These methods process and analyze images by breaking them down into smaller components known as image patches, finding prototypical parts, and combining evidence from the prototypes to make a final classification.

\subsubsubsection{Feature}  

Unlike methods which target entire images or image patches, local explanation methods focusing on features are concerned with the abstract representations that a model extracts from an image such as SpRAy~\cite{lapuschkin2019unmasking} and InterVENE~\cite{nauta2020interactive}.
These features, often in the form of embeddings or activations within hidden layers, help to understand the model’s behavior at a deeper level.

\subsubsection{Semi-local}  

\textbf{Semi-local} explanations occupy a position between purely local and global approaches.
Rather than focusing on individual samples or the entire model, Semi-local explanations target a group of samples that share common semantic concept or belong to the same category.
Semi-local explanations can mainly be categorized into the following types:

\subsubsubsection{Instance / Entity} 

Instance-level interpretability focuses on explaining how individual entities -- such as faces, objects, or persons -- are recognized by the model.
These methods aim to elucidate the model's behavior by explaining instances, which are commonly applied in domains such as face recognition and vehicle identification~\cite{yin2019towards,lin2021xcos}.

\subsubsubsection{Semantic Concept}

Explanation methods targeting semantic concepts focus on groups of samples that share common semantic meanings or features~\cite{kim2018interpretability,chen2019looks,koh2020concept,li2018deep,chen2020concept,zhou2018interpretable,yeh2020completeness,cheng2020explaining,hase2019interpretable,zhang2021invertible,sawada2022concept,sarkar2022framework,wang2023learning,ma2024looks,ramaswamy2022elude,posada2024eclad,ramaswamy2023ufo}.
These methods typically recognize objects through attributes or semantically meaningful concepts, highlighting the importance of accurately interpreting these concepts, such as \cite{chen2019looks}, \cite{koh2020concept}, and \cite{li2018deep}.
By analyzing these conceptually similar groups, the explanations can provide insights into how the model understands and processes these shared semantic elements.

\subsubsubsection{Category}   

These methods are designed for explaining a group of samples that belong to the same category~\cite{kim2018interpretability,bau2017network,hendricks2016generating,koh2020concept,ghorbani2019towards,zhou2018interpreting,yeh2020completeness,nauta2021neural,hendricks2018grounding,goyal2019explaining,zhou2018revisiting,sawada2022concept,achtibat2023attribution,nauta2023pip,schrouff2021best,hendricks2021generating,posada2024eclad,dreyer2023understanding,akata2018generating,hong2024concept,liu2019tabby, wan2024interpretable}, which is the most common scenario in the visual recognition tasks.
They focus on understanding the common features or patterns that the model uses to classify samples into the certain category.
In other words, the explanations delineate the categories from the model’s perspective, including the representative samples and the decision boundary.

\subsubsection{Global} 

In contrast to local and semi-local explanations, \textbf{global} explanations pertain to cases in which the entire model, rather than individual input samples, is the focus of explanation.
Global explanations generally offer a high-level overview of the model's working mechanism, which includes:

\subsubsubsection{Model}    

Model-level global explanations focus on understanding the overall architecture and behavior of the entire model.
This approach seeks to provide insights into how different components of the model, such as intermediate layers, contribute to its functioning and decision-making~\cite{zhou2016learning,bau2017network,zhou2018interpreting,zhou2018interpretable,dong2017towards,zhou2018revisiting,zhang2017growing,zhang2020interpretable,zhang2018unsupervised,shen2021interpretable,huang2022segdiscover,nauta2020interactive,peters2019visualising}.
For example, Network Dissection~\cite{bau2017network} quantifies the interpretability of CNNs by evaluating how hidden units align with human-interpretable semantic concepts.
Furthermore, decision rules similar to decision trees can be regarded as the direct interpretation of the model as well.

\subsubsubsection{Neuron}  

Neuron-level global explanations focus on the behavior and influence of individual neurons within the model~\cite{hernandez2021natural, peters2019visualising}.
These methods analyze how specific neurons or groups of neurons activate in response to different inputs and how their activations contribute to the overall output of the model.
For example, regarding to localization of neurons, \cite{zhou2018interpreting} revisits the role of individual units in CNNs by visualizing their activations using dimensionality reduction.
For semantic alignment of neurons, \cite{hernandez2021natural} automatically assigns natural language descriptions to neurons by leveraging mutual information, enabling open-ended and compositional interpretation of neuron functions.

\subsection{Presentation}
\label{sec:method-presentation}

\begin{figure}[t]
\centering
\includegraphics[width=\linewidth]{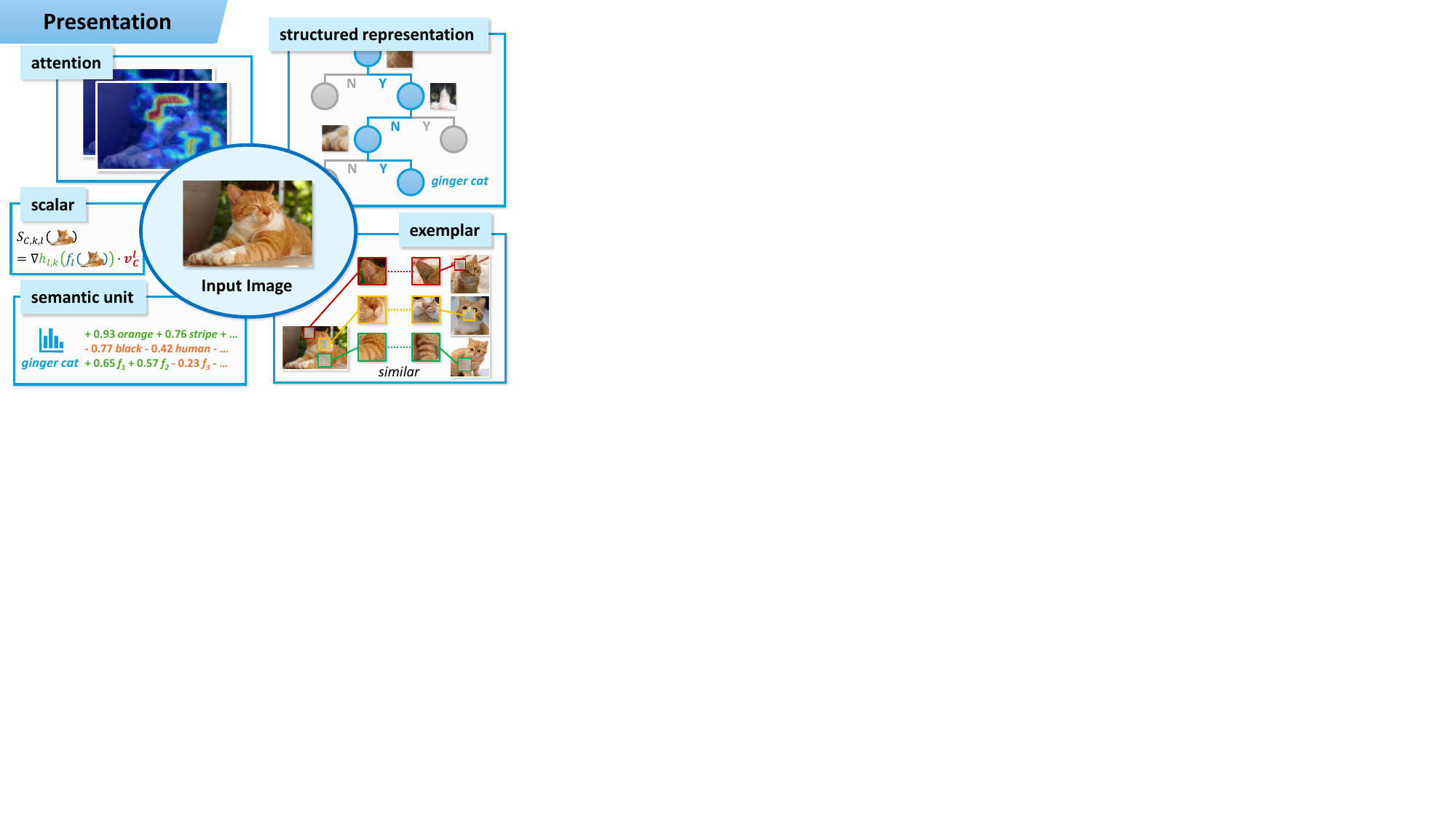}
\caption{Illustration of \textbf{Presentation}. Some representative examples for \textbf{scalar}~\cite{kim2018interpretability}, \textbf{attention}~\cite{selvaraju2017grad}, \textbf{structured representation}~\cite{nauta2021neural}, \textbf{semantic unit}~\cite{yeh2020completeness}, and \textbf{exemplar}~\cite{chen2019looks} are presented respectively.}
\label{fig:presentation}
\end{figure}

The \textbf{presentation} of interpretability refers to how explanations are expressed and perceived, and can be categorized according to the form of their outputs.
This dimension is particularly important for visual recognition models, as different presentation forms vary significantly in readability, intuitiveness, and suitability for different modalities.
Several surveys~\cite{schwalbe2024comprehensive,nauta2023anecdotal,seifert2017visualizations,borys2023explainable,saranya2023systematic} have highlighted presentation as a key axis for organizing XAI methods.
Considering the requirements of \textbf{localization interpretability} and \textbf{semantic interpretability}, explanations are commonly delivered through visual or textual modalities, and can be broadly grouped into several categories: \textbf{scalar}, \textbf{attention}, \textbf{structured representation}, \textbf{semantic unit}, and \textbf{exemplar}, as illustrated in Fig.~\ref{fig:presentation}.

\subsubsection{Scalar}

\textbf{Scalar} explanations present interpretability results as numerical scores, making them well suited for quantitative analysis and comparison.
Such outputs typically measure feature importance, concept relevance, or similarity through scalar values~\cite{kim2018interpretability,bau2017network,chen2019looks}.
For instance, TCAV~\cite{kim2018interpretability} quantifies the sensitivity of a model’s predictions to predefined concepts, while Network Dissection~\cite{bau2017network} evaluates the alignment between individual convolutional units and semantic concepts via unit-wise scores.
In prototype-based models such as ProtoPNet~\cite{chen2019looks}, scalar similarity scores between image regions and learned prototypes provide quantitative evidence supporting the final classification.

\subsubsection{Attention}

\textbf{Attention}-based explanations highlight the importance of different input regions through spatial masks, offering more intuitive visual cues than purely scalar outputs.
By directly localizing discriminative regions, attention-based presentations facilitate human understanding of model behavior and are widely adopted in visual recognition tasks.
Depending on the granularity of localization, attention-based outputs can be further divided into the following forms.

\subsubsubsection{Heatmap}    

Heatmaps are the most prevalent attention-based presentation, visualizing regional importance by overlaying color-coded intensity maps on the input image~\cite{selvaraju2017grad,zhou2016learning,sundararajan2017axiomatic,bach2015pixel}.
Classic attribution methods such as CAM~\cite{zhou2016learning}, Grad-CAM~\cite{selvaraju2017grad}, Integrated Gradients~\cite{sundararajan2017axiomatic}, LRP~\cite{bach2015pixel}, and SmoothGrad~\cite{smilkov2017smoothgrad} generate heatmaps by estimating each pixel’s contribution to the prediction.
Beyond gradient-based approaches, alternative strategies have also been explored: Concept-Centric Transformers~\cite{hong2024concept} visualize activations of learned latent concepts, while interpretability-aware ViTs~\cite{qiang2023interpretability} produce attention maps intrinsically aligned with discriminative patterns.

\subsubsubsection{Segmentation Mask}  

Segmentation masks provide a more precise form of attention by delineating semantically meaningful regions with clear boundaries~\cite{huang2022segdiscover,sun2024explain}.
Compared to heatmaps, they explicitly associate model decisions with object- or concept-level segments.
For example, SegDiscover~\cite{huang2022segdiscover} unsupervisedly extracts semantic visual concepts from complex scenes and visualizes them as segments in the latent space.
Explain Any Concept~\cite{sun2024explain} further integrates SAM~\cite{kirillov2023segment} with Shapley value estimation to generate segmentation masks that correspond to concepts critical for a model’s prediction.

\subsubsection{Structured Representation}

\textbf{Structured representations} convey explanations through explicit structures such as graphs or trees, enabling a more systematic depiction of the model’s reasoning process.
While these forms offer high interpretability, they typically require carefully defined nodes, edges, or decision rules, making them more challenging to construct.
Structured explanations are mainly instantiated in the following forms.

\subsubsubsection{Graph}

Graph-based explanations represent features, concepts, or components as nodes, with edges encoding relationships such as dependencies or causal interactions.
Interpretable Part Graphs~\cite{zhang2017growing}, for example, organize latent patterns mined from CNNs into a hierarchical and-or graph, revealing how object parts contribute to decisions.
Other works~\cite{li2018beyond,shi2019explainable} employ scene graphs as intermediate representations, enabling graph-based reasoning and providing naturally interpretable explanations aligned with visual structure.

\subsubsubsection{Tree}  

Tree-based explanations describe decision-making as a hierarchical process, where branches correspond to decision criteria and leaves represent outcomes.
Such representations are commonly used to explain predictions via explicit decision paths.
For instance, \cite{liu2019tabby} leverages category hierarchies to derive attribute-based decision rules, while ProtoTree~\cite{nauta2021neural} integrates prototype learning with decision trees to explain individual predictions.
InterVENE~\cite{nauta2020interactive} further employs decision trees to interpret neural embeddings and supports interactive exploration of neuron-level explanations.

\subsubsection{Semantic Unit}

Explanations with \textbf{semantic units} decompose predictions into semantically meaningful components, enabling concept-level understanding of model behavior.
Due to their emphasis on semantic interaction, these approaches are often closely associated with natural language representations.
They mainly fall into the following subcategories.

\subsubsubsection{Attribute / Concept}   

Concept-based explanations analyze how high-level semantic concepts influence model decisions~\cite{koh2020concept,chen2020concept,yeh2020completeness}.
The most representative framework is the Concept Bottleneck Model (CBM)~\cite{koh2020concept}, which predicts human-defined concepts before using them for final classification, thereby providing concept-level explanations at inference.
Subsequent works have extended CBMs in various directions~\cite{sawada2022concept,heidemann2023concept,lockhart2022learn,oikarinen2023label}, while concept discovery methods~\cite{ghorbani2019towards,yeh2020completeness,huang2022segdiscover} aim to automatically infer interpretable concept sets without requiring additional annotations.

\subsubsubsection{Description}

This subgroup focuses on generating natural language descriptions that explain neuron functionality or model attention~\cite{hendricks2016generating,park2018multimodal,hendricks2018grounding}.
Such descriptions provide direct, human-readable explanations that bridge visual evidence and semantic interpretation.
For example, MILAN~\cite{hernandez2021natural} produces textual descriptions for individual neurons, while pointing-and-justification approaches~\cite{park2018multimodal} combine attention localization with language-based explanations.

\subsubsection{Exemplar}

\textbf{Exemplar}-based explanations illustrate model behavior through representative visual examples, making them highly intuitive and accessible.
Rather than abstract scores or masks, exemplars expose the evidence used by the model in a concrete form.
They can be categorized into the following types.

\subsubsubsection{Embedding Visualization}

Embedding visualization methods project high-dimensional feature representations into low-dimensional spaces for human inspection.
InterVENE~\cite{nauta2020interactive} visualizes neural embeddings and supports interactive analysis, while tools such as~\cite{peters2019visualising} employ dimensionality reduction techniques to illustrate training dynamics and feature separability.
These visualizations help reveal feature organization and class structure within learned representations.

\subsubsubsection{Prototype}

Prototype-based explanations present representative examples that the model relies on for decision-making~\cite{chen2019looks,nauta2021neural}.
ProtoPNet~\cite{chen2019looks}, for instance, explains predictions by matching image regions to learned prototypical parts.
Extensions such as ProtoConcepts~\cite{ma2024looks} enhance interpretability by learning concept-level prototypes composed of multiple patches, yielding richer visual explanations.

\subsubsubsection{Patch}

Patch-based explanations highlight localized image regions that influence predictions~\cite{ghorbani2019towards,nauta2021neural}.
Many prototype-based methods visualize patches as prototypical evidence, while ACE~\cite{ghorbani2019towards} discovers visual concepts directly from image patches.
Other works~\cite{hendricks2018grounding} ground linguistic explanations to image patches, forming semantic correspondences between text and visual regions.

\subsubsubsection{Generated Image}

With advances in generative modeling, several methods generate synthetic images to visualize neurons or decision boundaries.
DISSECT~\cite{ghandeharioun2021dissect}, for example, produces images that maximize neuron activation, revealing learned patterns.
Counterfactual approaches~\cite{zhao2021generating,matsui2022counterfactual} further employ generated images to illustrate how minimal changes can alter predictions.
Among all presentation forms, they offer the most direct and expressive visual explanations.

\subsection{Methodology}
\label{sec:method-methodology}

\begin{figure}[t]
\centering
\includegraphics[width=\linewidth]{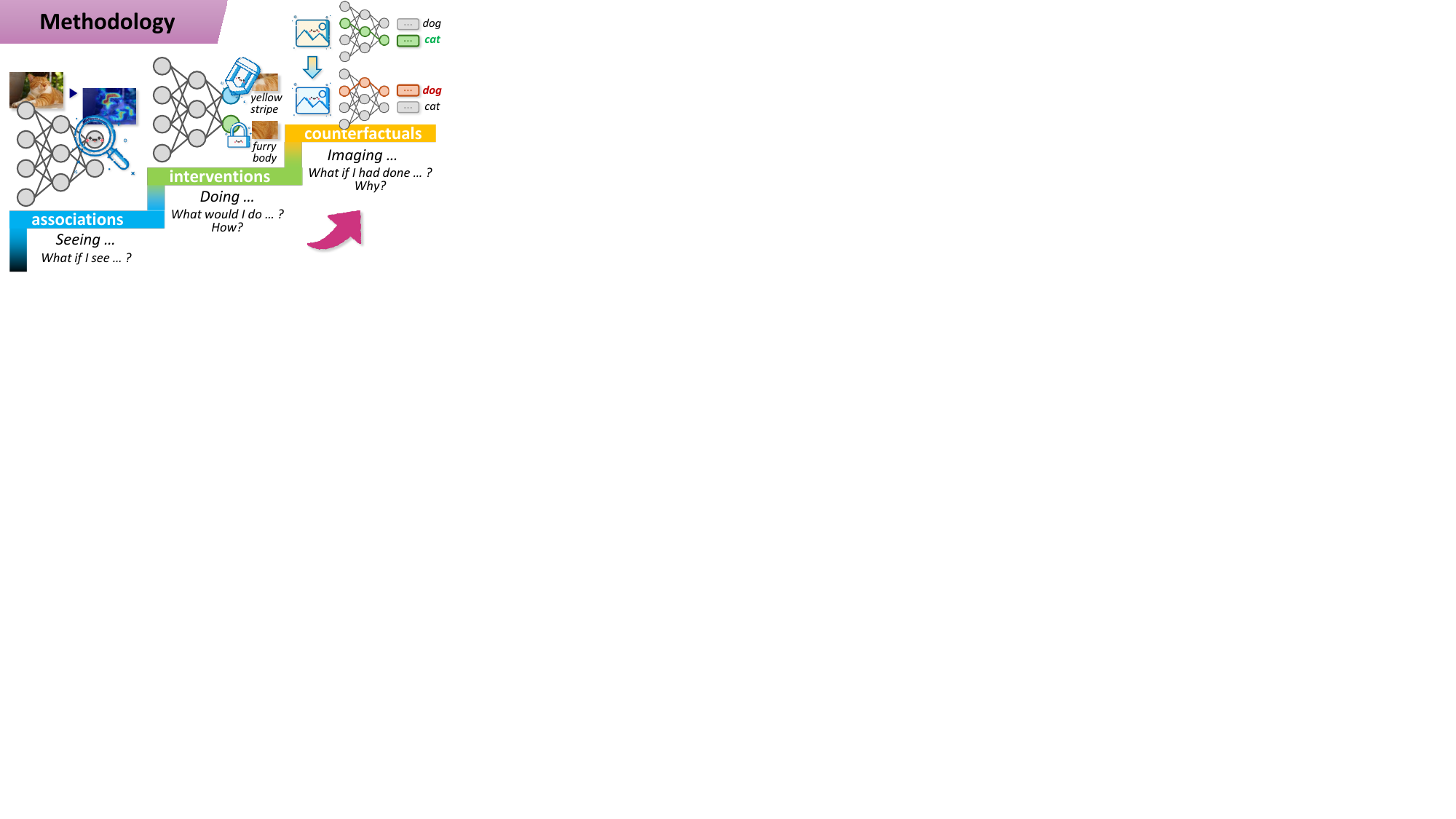}
\caption{Illustration of \textbf{Methodology}. Associations, interventions, and counterfactuals are three levels from the Ladder of Causation \cite{pearl2018book}.}
\label{fig:methodology}
\end{figure}

The interpretability \textbf{methodology} characterizes how explanations are derived, jointly determined by the intent of interpretability, the object being explained, and the representation of explanations.
Inspired by \textit{The Ladder of Causation}~\cite{pearl2018book}, we group methodologies into three paradigms: \textbf{association}, \textbf{intervention}, and \textbf{counterfactual}, which respectively explain models by discovering input--output relationships, manipulating internal mechanisms, and probing ``what-if'' alternatives.
Previous surveys \cite{nauta2023anecdotal,nazir2023survey,hohl2024opening,ras2022explainable,gilpin2018explaining} also categorize methods from methodology, and we propose a more detailed multi-layer categorization (Fig. \ref{fig:methodology}).

\subsubsection{Association}

\textbf{Association-based} methods explain model behavior by identifying statistical or functional relationships between signals, most commonly between inputs and outputs.
They often operate post hoc and thus cover a large portion of passive interpretability for visual recognition.
Depending on what relationship is extracted and how it is presented, association-based explanations can be instantiated as attribution, low-dimensional visualization, clustering, example-based evidence, generation, or distillation.

\subsubsubsection{Attribution}

Attribution methods assign importance scores to input features or intermediate components and typically visualize them as heatmaps for spatial localization~\cite{selvaraju2017grad,zhou2016learning,sundararajan2017axiomatic,bach2015pixel,smilkov2017smoothgrad}.
The CAM family is a representative line: CAM~\cite{zhou2016learning} uses the activation maps before global average pooling in CNNs to highlight discriminative regions, while Grad-CAM~\cite{selvaraju2017grad} incorporates gradients to generalize CAM-style explanations across a wider range of architectures.
LRP~\cite{bach2015pixel} propagates relevance scores backward layer by layer to trace how neurons contribute to the final prediction.
More recently, concept-oriented extensions such as Explain Any Concept~\cite{sun2024explain} build upon SAM~\cite{kirillov2023segment} to provide concept-level localization, connecting attribution with semantically grounded explanations.

\subsubsubsection{Dimensionality Reduction}  

Dimensionality reduction methods project high-dimensional embeddings into low-dimensional spaces to make global structure and class separability visually inspectable.
InterVENE~\cite{nauta2020interactive} exemplifies this direction by projecting neural embeddings into a 2D scatter plot and enabling interactive exploration for interpretation.
Another line of work~\cite{peters2019visualising} visualizes the evolution of neuronal activity across training epochs, producing trajectories that reflect how representations change over time.
Overall, these methods emphasize global geometry and dynamics of representations, complementing attribution methods that focus on localized evidence.

\subsubsubsection{Clustering}   

Clustering-based explanations group samples, features, or explanation maps according to similarity, revealing common decision patterns and potential biases.
SpRAy~\cite{lapuschkin2019unmasking} clusters relevance maps (e.g., produced by LRP~\cite{bach2015pixel}) to identify recurring prediction strategies and to expose spurious cues shared across instances.
ECLAD~\cite{posada2024eclad} instead clusters pixel-level activation maps to automatically extract and localize concepts from CNNs, turning latent activations into concept-like groupings.
By organizing explanations into clusters, these methods provide a higher-level view of how a model behaves across samples rather than explaining a single prediction in isolation.

\subsubsubsection{By Example}

Example-based methods explain predictions through representative instances, prototypes, or patches, linking model outputs to concrete evidence in the dataset~\cite{chen2019looks,li2018deep,nauta2021neural,nauta2023pip,wang2023learning,ma2024looks,wan2024interpretable}.
ProtoPNet~\cite{chen2019looks} is a canonical approach: it matches image parts to learned prototypes and aggregates such evidence for classification.
ProtoPool~\cite{rymarczyk2022interpretable} improves scalability by sharing prototypes across classes, reducing prototype redundancy and simplifying training.
ST-ProtoPNet~\cite{wang2023learning} focuses on prototypes near decision boundaries to improve predictive performance, while ProtoConcepts~\cite{ma2024looks} learns concept-level prototypes composed of multiple patches for richer visual evidence.
SPANet~\cite{wan2024interpretable} further associates prototypes with semantic labels, bridging example-based explanations with concept-level interpretability.

\subsubsubsection{Generation}  

Generation-based methods construct explanatory artifacts---typically text or images---to reveal what a model has learned or which patterns it is sensitive to~\cite{hendricks2016generating,akata2018generating,hendricks2018grounding,hendricks2021generating,ghandeharioun2021dissect}.
Generating Visual Explanations~\cite{hendricks2016generating,akata2018generating} produces sentence-level explanations by combining image evidence with class descriptions.
Grounding Visual Explanations~\cite{hendricks2018grounding,hendricks2021generating} adds a grounding mechanism to align textual phrases with localized visual regions, improving faithfulness and category discriminativeness.
Beyond language, DISSECT~\cite{ghandeharioun2021dissect} uses concept traversals to generate sequences of examples that illustrate how latent factors influence decision boundaries.
These methods provide interpretable artifacts that are directly consumable by humans, complementing score-based and example-based evidence.

\subsubsubsection{Knowledge Distillation}    

Knowledge distillation-based explanations transfer behavior from a complex teacher to a simpler, more interpretable student, enabling interpretation via the student’s structure.
ELUDE~\cite{ramaswamy2022elude} distills black-box predictions into an interpretable linear classifier built on concept labels, aiming to retain performance while improving transparency.
IA-ViT~\cite{qiang2023interpretability} distills knowledge from a ViT predictor to an interpreter module that reproduces predictive distributions and outputs attention maps, where the single-head self-attention serves as an explicit explanatory mechanism.
By replacing opaque computation with a constrained surrogate, distillation offers a pragmatic route to faithful explanations when the original model is difficult to interpret directly.

\subsubsection{Intervention}    

\textbf{Intervention}-based methods explain models by actively manipulating internal representations or training objectives, and then observing how such changes affect predictions.
Compared with association-based approaches, intervention provides more direct evidence of causal influence within the model.
These methods can be categorized by the target of intervention, such as rectifying feature activations or constraining semantics.

\subsubsubsection{Feature Rectification}   

Feature-rectification approaches introduce specialized losses or architectural constraints so that feature maps exhibit interpretable spatial behaviors~\cite{shen2021interpretable,zhang2020interpretable,dong2017towards}.
For instance, \cite{dong2017towards} encourages high-level filters to consistently encode specific object parts by suppressing responses that correspond to different semantic concepts.
Interpretable CNNs (ICNN)~\cite{zhang2018interpretable,zhang2020interpretable} apply losses on high-level convolutional feature maps to encourage each filter to represent a distinct object part, yielding interpretable convolutional layers.
Interpretable Compositional CNNs~\cite{shen2021interpretable} further relax shape assumptions by allowing filters to represent irregular parts or regions without clear structures, broadening applicability while maintaining part-based interpretability.

\subsubsubsection{Semantic Constraint}

Semantic-constraint methods enforce interpretability by aligning internal representations with human-understandable concepts~\cite{bau2017network,koh2020concept,chen2020concept,ramaswamy2023ufo}.
Network Dissection~\cite{bau2017network} evaluates interpretability by measuring how well neurons align with semantic concepts (e.g., objects, parts, textures, scenes), providing a concept-level lens for understanding learned representations.
Concept Bottleneck Models (CBMs)~\cite{koh2020concept} impose a stronger constraint by requiring the model to predict human-defined concepts first and then use them for classification, producing concept-based explanations by design.
Subsequent variants such as CBM-AUC~\cite{sawada2022concept} combine supervised and discovered concepts, often expanding the bottleneck dimensionality to mitigate incompleteness of human-defined concept sets while preserving semantic alignment.

\subsubsection{Counterfactual}  

\textbf{Counterfactual-based} methods explain predictions through ``what-if'' reasoning: they alter inputs or conditions and examine how outputs change.
Such explanations highlight decisive factors by contrasting the original prediction with plausible alternatives.
Depending on whether modifications are simple edits or model-based synthesis, counterfactual methods can be instantiated as perturbation or generative counterfactuals.

\subsubsubsection{Perturbation}

Perturbation-based methods deliberately modify inputs and measure output changes to identify influential features or concepts~\cite{goyal2019explaining,nauta2021looks,zhou2018revisiting,fel2023don}.
A key advantage is model-agnosticism, which makes perturbation useful for black-box settings.
Explaining Prototypes~\cite{nauta2021looks} performs controlled edits (e.g., hue, texture, shape, contrast, saturation) and evaluates how similarity to prototypes changes, revealing which visual attributes drive recognition.
CaCE~\cite{goyal2019explaining} instead uses conditional generative models to construct counterfactual examples and estimate the causal effect of concept presence or absence on the classifier output, thus connecting perturbation with concept-level causal analysis.

\subsubsubsection{Generative Counterfactual}

While perturbations typically seek minimal edits, generative counterfactuals synthesize more substantial yet plausible alternatives using generative models such as GANs or diffusion models~\cite{van2021conditional,nemirovsky2022countergan,khorram2022cycle,mertes2022ganterfactual,pegios2024diffusion}.
These methods can modify attributes, structure, or regions in a controlled manner, producing counterfactual evidence that is closer to the data manifold.
In medical imaging, patch modification has been shown effective for probing model reliance on localized cues~\cite{li2024contrastive}, where contrasting positive and negative counterfactual samples helps reveal robust explanatory factors.
Overall, generative counterfactuals complement simple perturbations by trading off minimality for realism and expressiveness in the generated explanations.

\subsection{Case Study: Applying the Taxonomy}
\label{sec:method-case}

\begin{figure}[t]
\centering
\includegraphics[width=\linewidth]{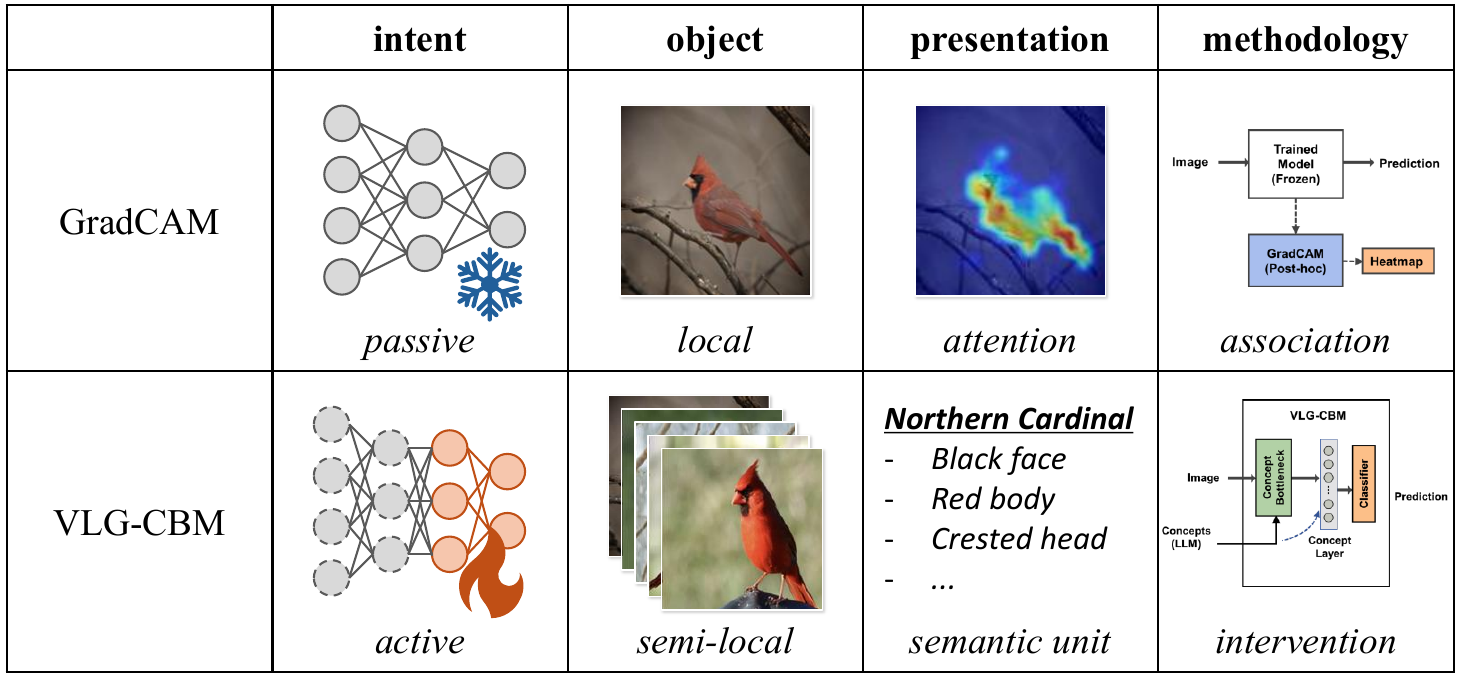}
\caption{Illustration of the proposed four-dimensional taxonomy applied to Grad-CAM~\cite{selvaraju2017grad} and VLG-CBM~\cite{srivastava2024vlg}.}
\label{fig:case}
\end{figure}

To demonstrate the usability and generality of the proposed four-dimensional taxonomy, we analyze two representative explainable visual recognition methods as examples: Grad-CAM~\cite{selvaraju2017grad}, a foundational cornerstone of post-hoc attribution, and VLG-CBM~\cite{srivastava2024vlg}, a state-of-the-art paradigm in inherently interpretable modeling (Fig.~\ref{fig:case}).

Grad-CAM~\cite{selvaraju2017grad} exemplifies the \textbf{passive} explainability paradigm. As a post-hoc method, it generates explanations for a pre-trained model without altering its inference logic.
It provides \textbf{local} explanations by highlighting spatial importance through \textbf{attention-based} heatmaps.
Methodologically, it relies on \textbf{association}, utilizing gradient information to attribute model decisions to specific input regions.
In contrast, VLG-CBM~\cite{srivastava2024vlg} represents a shift toward \textbf{active} explainability.
By integrating a vision-language concept bottleneck layer into the architecture, it ensures that the model’s reasoning is intrinsically tied to human-understandable \textbf{semantic units}.
This approach offers \textbf{semi-local} explanations that capture shared category-level patterns rather than sample-specific noise.
Methodologically, it operates through \textbf{intervention}, imposing structural constraints during training to guide the model's decision-making.

The mapping of these representative methods, ranging from classical attribution to modern interventionist frameworks, illustrates the broad applicability of our unified taxonomy across distinct stages of XAI development.
Furthermore, as the taxonomy is designed based on the universal architecture of recognition models and fundamental user requirements, it maintains long-term adaptability and extensibility, allowing it to accommodate future novel methods by expanding specific groups rather than modifying the core four-dimensional structure.

\begin{table*}[t]
\centering
\caption{Comparative overview of cross-dimensional XAI methods in visual recognition}
\label{tab:taxonomy-evaluation}
\begin{tabularx}{\linewidth}{lll|P{0.8cm}P{0.8cm}|CC|P{2cm}}
\toprule
\textbf{Dimension} & \textbf{Group} & \textbf{Description} & \textbf{Loc.} & \textbf{Sem.} & \textbf{Undst.} & \textbf{Fidel.} & \textbf{Practical Use} \\
\midrule
\multirow{2}{*}{Intent} & Passive & Explaining a pre-trained model & $\checkmark$ & $\checkmark$ & \starY{2} & \starY{1} & Easy \\
 & Active & Built-in interpretability during design & $\checkmark$ & $\checkmark$ & \starY{3} & \starY{3} & Hard \\
\midrule
\multirow{3}{*}{Object} & Local & Sample-level (individual) explanation & $\checkmark$ & $\checkmark$ & \starY{3} & \starY{2} & Easy \\
 & Semi-local & Explanation for a cluster/class & & $\checkmark$ & \starY{2} & \starY{2} & Moderate \\
 & Global & Entire model decision rule distillation & & $\checkmark$ & \starY{1} & \starY{3} & Hard \\
\midrule
\multirow{5}{*}{Presentation} & Scalar & Numerical importance scores & $\checkmark$ & $\checkmark$ & \starY{1} & \starY{3} & Easy \\
 & Attention & Saliency masks/heatmaps & $\checkmark$ & & \starY{2} & \starY{3} & Easy \\
 & Structured & Logical forms like trees or graphs & $\checkmark$ & $\checkmark$ & \starY{3} & \starY{2} & Hard \\
 & Semantic Unit & Decomposing into human concepts & & $\checkmark$ & \starY{2} & \starY{3} & Moderate \\
 & Exemplar & Using typical examples as analogy & $\checkmark$ & $\checkmark$ & \starY{3} & \starY{1} & Moderate \\
\midrule
\multirow{3}{*}{Methodology} & Association & Statistical correlations (Passive) & $\checkmark$ & $\checkmark$ & \starY{2} & \starY{1} & Easy \\
 & Intervention & Causal effects via active changes & $\checkmark$ & $\checkmark$ & \starY{3} & \starY{3} & Moderate \\
 & Counterfactual & Perturbation-based ``what-if'' scenarios & $\checkmark$ & $\checkmark$ & \starY{3} & \starY{2} & Easy \\
\bottomrule
\end{tabularx}%
\end{table*}

\subsection{Comparison and Summary}
\label{sec:method-summary}

To provide a structured overview of XAI in visual recognition, we qualitatively compare representative groups across key aspects (Tab.~\ref{tab:taxonomy-evaluation}).
This analysis captures empirical tendencies rather than quantitative measurements, characterizing each group by its suitability for \textbf{Localization} (\textbf{Loc.}) and \textbf{Semantic} (\textbf{Sem.}) \textbf{Interpretability}, and two cross-dimensional indicators: \textbf{Understandability} (\textbf{Undst.}), reflecting how intuitively humans grasp the explanations; \textbf{Fidelity} (\textbf{Fidel.}), reflecting the alignment between explanations and the model’s internal logic.
Additionally, we examine the \textbf{Practical Use} of each group to reflect its overall ease of implementation.

In the \textbf{intent} dimension, \textbf{passive} methods prioritize \textit{Practical Use} as they apply to any pre-trained model without modification, though they typically offer moderate \textbf{understandability} and limited \textbf{fidelity}.
Conversely, \textbf{active} methods achieve superior \textbf{fidelity} and \textbf{understandability} by embedding explanations directly into the decision logic, albeit at the cost of \textbf{practical use} due to their reliance on model-specific architectures and training.

Regarding the \textbf{object} dimension, \textbf{local} explanations offer intuitive, instance-level insights and are easy to apply.
\textbf{Global} methods aim for the highest \textbf{fidelity} by capturing holistic decision rules, albeit with lower \textbf{understandability} due to their abstraction.
\textbf{Semi-local} approaches strike a balance, providing moderate \textbf{understandability}, \textbf{fidelity}, and ease of implementation by explaining sample groups.

The \textbf{presentation} dimension reflects trade-offs between expressive power and implementation effort.
\textbf{Scalar} and \textbf{attention-based} presentations maintain high \textbf{fidelity} by directly reflecting internal importance signals, and are generally easy to apply due to their non-intrusive extraction.
\textbf{Structured} and \textbf{exemplar-based} presentations prioritize human intuition, yielding superior \textbf{understandability} despite higher implementation complexity.
\textbf{Semantic units} effectively combines semantic interpretability with robust \textbf{fidelity}, while requiring moderate effort to construct.

Finally, the \textbf{methodology} dimension reflects varying levels of causal engagement.
\textbf{Association-based} methods are the most accessible (\textbf{practical use}) but rely on correlations that may limit \textbf{fidelity}.
\textbf{Intervention-based} approaches improve \textbf{fidelity} through active manipulation, while \textbf{counterfactual} methods provide the most intuitive ``what-if'' insights, though at a computational or generative cost.

Overall, no single category dominates all aspects. Each reflects distinct trade-offs, and our taxonomy serves as a framework for informed method selection based on specific interpretability needs and practical constraints.

\section{Metrics}
\label{sec:metric}

Unlike standard visual recognition, evaluating interpretability is inherently subjective as it centers on human understanding.
While user studies are often seen as the gold standard, they are costly, prone to bias, and difficult to scale or compare.
Consequently, establishing robust quantitative metrics remains a significant challenge.
Despite the lack of a universal consensus, researchers have proposed various task-specific metrics and defined key properties for evaluation.
This section first outlines the fundamental requirements for interpretability evaluation in Sec.~\ref{sec:metric-req}, followed by a review of existing quantitative metrics in Sec.~\ref{sec:metric-list}.
In Sec.~\ref{sec:metric-pracital}, we conduct experiments to demonstrate how different metrics reflect specific desiderata and how they are used for benchmarking methods within specific sub-fields.

\subsection{Desiderata of Metrics}
\label{sec:metric-req}

While terminology varies across the literature, many proposed requirements for interpretability overlap in meaning.
We consolidate these criteria, and first focus on the core desiderata essential to visual XAI as listed below:

\begin{itemize}[leftmargin=*]

\item \textbf{Understandability}: The degree to which humans can comprehend an explanation \cite{samek2023explainable,ramaswamy2023ufo}. It is conceptually related to \textit{Meaningfulness} \cite{ghorbani2019towards}, \textit{Conciseness} \cite{zhou2018interpretable}, \textit{Interpretability} \cite{gilpin2018explaining,abusitta2024survey}, and qualities like \textit{Compactness}, \textit{Coherence}, and \textit{Complexity} \cite{nauta2023co,hedstrom2023quantus}.

\item \textbf{Fidelity}: The extent to which an explanation faithfully reflects model's behavior~\cite{samek2023explainable,hedstrom2023quantus}, including~\cite{dembinsky2025unifying}: 
(1) \textit{Correctness}, measuring the relevance of highlighted factors~\cite{nauta2023co}, also termed \textit{Importance}~\cite{ghorbani2019towards}, \textit{Objectiveness}~\cite{zhang2019towards}, or \textit{Generalizability}~\cite{fel2022good}; 
(2) \textit{Completeness}, assessing if all decision-relevant factors are captured~\cite{gilpin2018explaining,nauta2023anecdotal}.

\item \textbf{Continuity}: The cohesion between similar explanations and the discriminability of different ones. It involves structural properties such as \textit{Coherency} \cite{ghorbani2019towards}, \textit{Distinctness} \cite{ghandeharioun2021dissect}, and \textit{Separability} \cite{das2020opportunities}. Additionally, it denotes the resilience to perturbations \cite{samek2023explainable,nauta2023co}, often referred to as \textit{Stability} \cite{das2020opportunities}, \textit{Robustness} \cite{hedstrom2023quantus,zhang2019towards}, or \textit{Consistency} \cite{fel2022good}.

\item \textbf{Efficiency}: The computational resources and time required for explanation generation \cite{carvalho2019machine,dembinsky2025unifying}, critical for practical deployment. It is also characterized by its \textit{Complexity} \cite{robnik2018perturbation,bommer2024finding,klein2024navigating} or \textit{Runtime}~\cite{samek2023explainable}.

\end{itemize}

These core desiderata represent universal priorities across the visual XAI landscape.
\textbf{Understandability} and \textbf{Fidelity} exhibit significant cross-dimensional differences, allowing us to leverage them for a qualitative comparison of different interpretability method groups in Sec.~\ref{sec:method-summary}.
In contrast, \textbf{Continuity} and \textbf{Efficiency} are more closely tied to the technical implementation details of specific methods.
Other desiderata proposed in previous work are not discussed here, as they are either tailored to specific tasks or have not yet been systematically organized, such as \textit{Applicability}~\cite{samek2023explainable}, \textit{Implementation Constraints}~\cite{das2020opportunities}, \textit{Controllability}~\cite{nauta2023anecdotal,nauta2023co}, \textit{Utility} and \textit{Usability}~\cite{poeta2023concept,abusitta2024survey}.

Various surveys have attempted to categorize these diverse properties from different perspectives.
For instance, \cite{nauta2023co} proposes the Co-12 properties across three levels: \textit{Content}, \textit{Presentation}, and \textit{User}.
Others classify metrics based on the degree of human involvement, distinguishing between \textit{human-grounded} and \textit{functionally-grounded} types \cite{schwalbe2024comprehensive}, or more broadly, between \textit{human-centered} and \textit{computer-centered} categories \cite{mersha2024explainable}.
Notably, a recent and comprehensive review in~\cite{dembinsky2025unifying} synthesizes the fragmented landscape of XAI evaluation, providing a holistic perspective on how explainability assessment techniques have evolved.
By consolidating these desiderata, we aim to encapsulate the community's expectations and provide a structured guide for future research on evaluating visual XAI methods.

\subsection{Existing Metrics}
\label{sec:metric-list}

\begin{table}[t]
\centering
\begin{threeparttable}
\caption{Existing visual XAI metrics categorized by Desiderata.}
\label{tab:metrics}
\renewcommand{\arraystretch}{1.1}
\begin{tabularx}{\linewidth}{lCCCC}
\toprule
Metric & \textbf{Undst.} & \textbf{Fidel.} & \textbf{Conti.} & \textbf{Effic.} \\
\midrule
\multicolumn{5}{l}{\textbf{Localization Metrics}} \\
\quad AOPC~\cite{samek2016evaluating} & \ding{51} & \ding{51} & & \\
\quad Pointing Game (PG)~\cite{zhang2018top} & & \ding{51} & \ding{51} & \\
\quad Deletion, Insertion~\cite{petsiuk2018rise} & & \ding{51} & \ding{51} & \\
\quad MCS, IDR~\cite{yang2019benchmarking} & & \ding{51} & & \\
\quad IIR~\cite{yang2019benchmarking} & & & \ding{51} & \\
\quad Bias of Attribution Map~\cite{zhang2018top} & & \ding{51} & & \\
\quad Unexplainable Feature~\cite{zhang2018top} & & \ding{51} & & \\
\quad Robustness, Mutual Verif.~\cite{zhang2018top} & & & \ding{51} & \\
\quad Faithfulness ($F$)~\cite{tomsett2020sanity} & & \ding{51} & & \\
\quad HI score~\cite{theodorus2020evaluating} & \ding{51} & \ding{51} & \ding{51} & \\
\quad POMPOM~\cite{rio2020understanding} & & \ding{51} & \ding{51} & \\
\quad FP Error, FN Error~\cite{mohseni2021quantitative} & & \ding{51} & & \\
\quad iAUC, IntIoSR~\cite{li2021experimental} & & \ding{51} & & \\
\quad CS, SENSmax~\cite{li2021experimental} & & & \ding{51} & \\
\quad GTC, SC, IoU~\cite{boggust2022shared} & & \ding{51} & & \\
\quad MeGe, ReCo~\cite{fel2022good} & & \ding{51} & \ding{51} & \\
\quad RMA, RRA~\cite{arras2022clevr} & & \ding{51} & & \\
\quad AR, AP~\cite{hagos2023distance} & & \ding{51} & & \\
\addlinespace[0.1em]
\multicolumn{5}{l}{\textbf{Semantic Metrics}} \\
\quad Completeness Score~\cite{yeh2020completeness} & & \ding{51} & & \\
\quad $N^{fg}_{concept}$, $N^{bg}_{concept}$, $\lambda$ ratio~\cite{cheng2020explaining} & & \ding{51} & & \\
\quad $\rho$ value~\cite{cheng2020explaining} & & & \ding{51} & \\
\quad $D_{mean}$, $D_{std}$~\cite{cheng2020explaining} & & & & \ding{51} \\
\quad $Fid_c$, $Fid_r$~\cite{zhang2021invertible} & & \ding{51} & & \\
\quad Faithfulness, Fidelity~\cite{sarkar2022framework} & & \ding{51} & & \\
\quad Explanation Error~\cite{sarkar2022framework} & \ding{51} & \ding{51} & & \\
\quad Intervention on Concepts (IoC)~\cite{sarkar2022framework} & & \ding{51} & & \\
\quad AIPD, AIFD~\cite{wang2023learning} & & & \ding{51} & \\
\quad Factuality, Groundability~\cite{yang2023language} & \ding{51} & \ding{51} & & \\
\quad CDR~\cite{wang2023learning2} & \ding{51} & & & \\
\quad CC, MIC~\cite{wang2023learning2} & & & \ding{51} & \\
\quad TCPC, TOPC~\cite{lai2023faithful} & & & \ding{51} & \\
\quad CUE~\cite{shang2024incremental} & \ding{51} & & & \ding{51} \\
\quad RC, IC~\cite{posada2024eclad} & \ding{51} & \ding{51} & & \\
\bottomrule
\end{tabularx}
\begin{tablenotes}
    \footnotesize
    \item Full table available at: \url{https://vipl-vsu.github.io/xai-recognition/}.
\end{tablenotes}
\end{threeparttable}
\end{table}

\begin{figure}[t]
\centering
\includegraphics[width=0.5\textwidth]{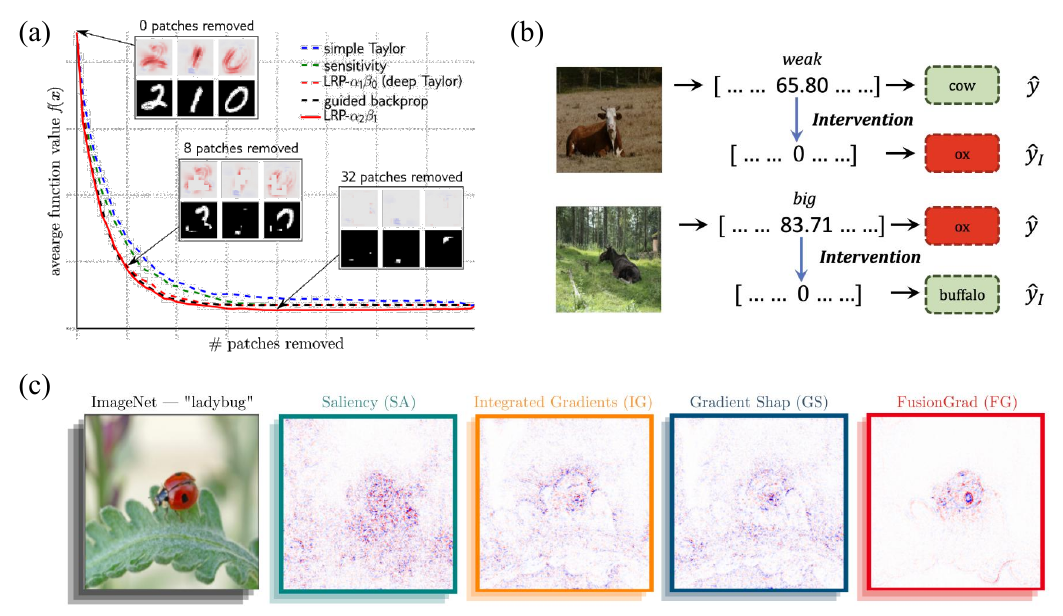}
\caption{Examples on interpretability evaluation. (a) An occlusion process can be employed to evaluate the quality of explanations~\cite{montavon2018methods}, which has been utilized in localization metrics such as AOPC~\cite{samek2016evaluating}. (b) Concept intervention always highlights the importance of concepts in decision-making and is frequently employed as a semantic metric in concept-based methods~\cite{sarkar2022framework}. (c) The Quantus toolkit is used to assess the interpretability of various attribution methods~\cite{hedstrom2023quantus}.}
\label{fig:metric}
\end{figure}

For quantitative evaluation, researchers have proposed some proxy metrics for interpretability in visual recognition.
However, these metrics are often constrained by the specific characteristics of the tasks to which they are applied, limiting their universal applicability.
As discussed in Sec.~\ref{sec:taxonomy}, based on the modalities that XAI methods target, mainstream research directions can be classified into \textbf{localization interpretability} (pertaining to the visual modality) and \textbf{semantic interpretability} (pertaining to the textual modality).
Accordingly, within the context of interpretability evaluation metrics for recognition tasks, these metrics are typically divided into \textbf{localization metrics} and \textbf{semantic metrics}.
We organize them according to the \textbf{desiderata} they validate, and examples of these metrics are presented in Tab.~\ref{tab:metrics}, where \textbf{Undst.}, \textbf{Fidel.}, \textbf{Conti.}, and \textbf{Effic.} denote \textbf{Understandability}, \textbf{Fidelity}, \textbf{Continuity}, and \textbf{Efficiency}, respectively.
Fig.~\ref{fig:metric} illustrates the discussed evaluation procedures.

\subsubsection{Localization Metrics}

Localization metrics evaluate how accurately an explanation identifies the input regions responsible for a model's prediction. 
As shown in Tab.~\ref{tab:metrics}, \textbf{Fidelity} is the most widely validated desideratum for localization metrics. 
Metrics such as \textit{AOPC} \cite{samek2016evaluating}, \textit{Deletion} \cite{petsiuk2018rise}, and \textit{Insertion} \cite{petsiuk2018rise} measure fidelity through causal interventions by perturbing or removing salient pixels. 
Meanwhile, \textit{Pointing Game (PG)} \cite{zhang2018top}, \textit{IoU} \cite{boggust2022shared}, \textit{GTC} \cite{boggust2022shared}, and \textit{SC} \cite{boggust2022shared} quantify fidelity via spatial alignment with ground-truth masks. 
\textbf{Continuity} is also frequently assessed, with metrics like \textit{SENSmax} \cite{li2021experimental} and \textit{Robustness} \cite{zhang2018top} testing the stability of visual explanations under input perturbations. 
In contrast, \textbf{Understandability} is less commonly quantified in localization, appearing primarily in metrics like \textit{HI score} \cite{theodorus2020evaluating} that penalize visually fragmented or incoherent heatmaps. 
Notably, for localization methods, \textbf{Efficiency} can be measured through the generation time of explanations (e.g., heatmaps), which is a straightforward and practical evaluation approach.

\subsubsection{Semantic Metrics}

Semantic metrics assess explanations that rely on high-level concepts or textual descriptors. 
The distribution of desiderata in Tab.~\ref{tab:metrics} shows that semantic metrics place a higher emphasis on \textbf{Understandability} compared to localization metrics. 
Metrics such as \textit{Factuality} \cite{yang2023language}, \textit{Groundability} \cite{yang2023language}, and \textit{CDR} \cite{wang2023learning2} explicitly evaluate whether the discovered concepts are human-interpretable and factually accurate. 
\textbf{Fidelity} remains a core requirement, validated by metrics like the \textit{Completeness Score} \cite{yeh2020completeness} and \textit{Fid} \cite{zhang2021invertible, sarkar2022framework}, which check if the concept-based representation can faithfully reconstruct the model's decision. 
\textbf{Continuity} is measured by metrics such as \textit{TCPC} \cite{lai2023faithful} and \textit{TOPC} \cite{lai2023faithful}, focusing on the stability of concept weights. 
Finally, \textbf{Efficiency} in the semantic domain is quantified by metrics like \textit{CUE} \cite{shang2024incremental}, which considers the quantity and length of concepts, or $D_{mean}$ \cite{cheng2020explaining}, which tracks concept learning speed.

\subsubsection{Toolkits}

Several toolkits provide unified frameworks for comparing interpretability methods. 
\textit{Captum}~\cite{kokhlikyan2020captum} and \textit{Xplique}~\cite{fel2022xplique} implement a wide range of attribution and concept-based algorithms for PyTorch and TensorFlow, respectively. 
For standardized benchmarking, \textit{XAI-Bench}~\cite{liu2021synthetic} and \textit{Saliency-Bench}~\cite{zhang2023xai} offer curated datasets and rigorous metrics tailored for feature attribution and visual saliency. 
Furthermore, \textit{Quantus}~\cite{hedstrom2023quantus} holistically organizes evaluation approaches across dimensions such as faithfulness, robustness, and complexity.
For large vision-language models, LVLM Interpret~\cite{ben2024lvlm} provides an interactive interface to visualize image patch contributions and explore model behavior.
We observe that most existing toolkits primarily focus on localization metrics. 
This likely stems from the historical predominance of attribution-based methods in visual XAI, which has led to mature standardization in their evaluation.

\subsection{Quantitative Comparison}
\label{sec:metric-pracital}

\begin{table}[t]
\centering
\caption{Comparison of visual XAI methods for localization interpretability.}
\label{tab:metric-localization}
\begin{tabularx}{\linewidth}{lCCCC}
\toprule
\textbf{Desiderata} & \textbf{Undst.} & \textbf{Fidel.} & \textbf{Conti.} & \textbf{Effic.} \\
\textit{Metric} & \footnotesize\makebox[0pt]{\textit{$L_0$ norm} $\downarrow$} & \footnotesize\makebox[0pt]{\textit{Deletion} $\uparrow$} & \footnotesize\makebox[0pt]{\textit{Perturb.} $\downarrow$} & \footnotesize\makebox[0pt]{\textit{Runtime} $\downarrow$} \\
\midrule
GradCAM~\cite{selvaraju2017grad}      & 0.2295 & \textbf{0.1769} & 0.1324 & 108 ms \\
GradCAM++~\cite{chattopadhay2018grad}   & 0.2805 & 0.1624 & 0.0987 & 119 ms \\
% FullGrad~\cite{srinivas2019full}     & TODO   & TODO   & TODO   & TODO   \\
AblationCAM~\cite{desai_2020_WACV}  & 0.2616 & 0.1751 & 0.1153 & 3844 ms \\
FEM~\cite{fuad2020features}     & 0.2723 & 0.1196 & 0.0903 & \textbf{33 ms} \\
HiResCAM~\cite{draelos2020use}     & 0.2295 & 0.1769 & 0.1324 & 83 ms  \\
ScoreCAM~\cite{wang2020score}     & 0.3407 & 0.1584 & 0.0992 & 4814 ms   \\
XGradCAM~\cite{fu2020axiom}     & 0.2295 & 0.1769 & 0.1324 & 84 ms  \\
EigenCAM~\cite{muhammad2020eigen}     & 0.2990 & 0.0876 & 0.1723 & 887 ms \\
% EigenGradCAM~\cite{duguaeșescu2022evaluation} & 0.2523 & 0.0923 & 0.2175 & 314 ms \\
LayerCAM~\cite{jiang2021layercam}     & 0.2826 & 0.1564 & 0.0961 & 90 ms  \\
% GradCAM EW~\cite{sun2021getam}   & 0.2826 & 0.1564 & 0.0961 & 90 ms  \\
KPCA CAM~\cite{karmani2024kpca}     & 0.1328 & 0.1272 & \textbf{0.0787} & 36 ms  \\
ShapleyCAM~\cite{cai2025cams}   & 0.2295 & 0.1769 & 0.1324 & 67 ms  \\
FinerCAM~\cite{zhang2025finer}     & \textbf{0.1038} & 0.1660 & 0.1772 & 74 ms  \\
\bottomrule
\end{tabularx}
\end{table}

\begin{table}[t]
\centering
\caption{Comparison of visual XAI methods for semantic interpretability.}
\label{tab:metric-semantic}
\begin{tabularx}{\linewidth}{lCCCC}
\toprule
\textbf{Desiderata} & \textbf{Undst.} & \textbf{Fidel.} & \textbf{Conti.} & \textbf{Effic.} \\
\textit{Metric} & \footnotesize\makebox[0pt]{\textit{Concept\%} $\downarrow$} & \footnotesize\makebox[0pt]{\textit{IoC} $\uparrow$} & \footnotesize\makebox[0pt]{\textit{Perturb.} $\downarrow$} & \footnotesize\makebox[0pt]{\textit{Rel. Cost} $\downarrow$} \\
\midrule
CBM~\cite{koh2020concept}         & \textbf{0.2134} & 0.0154 & 0.0158 & +33.47\% \\
CBM-AUC~\cite{sawada2022concept}     & 0.2686 & 0.0268 & 0.0272 & +67.56\% \\
LaBo~\cite{yang2023language}        & 0.4842 & 0.0000 & 0.0111 & +12.37\% \\
LFCBM~\cite{oikarinen2023label}       & 0.4633 & 0.3036 & 0.0404 & +73.06\% \\
% CDL~\cite{zang2024pre}         & TODO   & TODO   & TODO   & TODO   \\
CSS VL-CBM~\cite{muthuchamy2024improving}  & 0.4194 & 0.2409 & 0.0580 & \textbf{+1.28\%} \\
VLG-CBM~\cite{srivastava2024vlg}     & 0.4339 & \textbf{0.8445} & \textbf{0.0021} & +5.77\% \\
\bottomrule
\end{tabularx}
\end{table}

Although the previously discussed metrics are mostly task-specific, they share a common foundation rooted in the core desiderata.
This convergence allows for quantitative benchmarking within specific sub-fields where explanation formats are comparable.
To demonstrate the practical application of these desiderata and metrics, we evaluate representative methods for both localization and semantic interpretability on the CUB dataset~\cite{wah2011caltech}, which is the most widely adopted benchmark in Visual XAI due to its comprehensive part-level and attribute-level annotations.
The evaluation is conducted using unified metrics or their variants.
For \textbf{localization interpretability}, \texttt{pytorch-grad-cam}~\cite{jacobgilpytorchcam} is used for implementation, and we operationalize the core desiderata as following metrics:

\begin{itemize}[leftmargin=*]
\item \textit{$L_0$ norm} (\textbf{Understandability}): The ratio of pixels with activation values above a threshold to the total image pixels. A lower value indicates a more concise and human-readable explanation.
\item \textit{Deletion} (\textbf{Fidelity}): The decrease in model confidence when the most important pixels (as ranked by the heatmap) are set to zero. A higher value suggests the method faithfully identifies decision-critical regions.
\item \textit{Perturb.} (\textbf{Continuity}): \textit{Perturbation Error}, the change in the heatmap when the input image undergoes data augmentation. Lower values indicate higher stability and resilience to input noise.
\item \textit{Runtime} (\textbf{Efficiency}): The average time required to generate a single explanation, where lower values indicate better suitability for real-time deployment.
\end{itemize}

For \textbf{semantic interpretability}, the desiderata are measured through the following metrics:

\begin{itemize}[leftmargin=*]
\item \textit{Concept\%} (\textbf{Understandability}): The percentage of concepts with activation levels above a threshold relative to the total concept pool. Lower percentages reflect higher sparsity and better conciseness.
\item \textit{IoC} (\textbf{Fidelity}): \textit{Intervention on Concept}, the score measured by zeroing out the most influential concepts and observing the drop in prediction confidence. Higher values indicate higher fidelity.
\item \textit{Perturb.} (\textbf{Continuity}): \textit{Perturbation Error}, the mean variation in the concept importance vector under input image augmentations. Lower values represent more consistent semantic explanations.
\item \textit{Rel. Cost} (\textbf{Efficiency}): \textit{Relative Cost}, defined as the percentage increase in inference time of the interpretable model relative to the black-box baseline. Values closer to $0\%$ indicate negligible computational overhead.
\end{itemize}

The quantitative results are summarized in Tab.~\ref{tab:metric-localization} and Tab.~\ref{tab:metric-semantic}.
The results highlight the inherent trade-offs between different desiderata.
In localization (Tab.~\ref{tab:metric-localization}), methods that achieve superior \textbf{Understandability} do not always lead in \textbf{Fidelity} or \textbf{Continuity}, suggesting that extreme conciseness may sometimes overlook secondary but relevant features.
In the semantic domain (Tab.~\ref{tab:metric-semantic}), while some methods achieve high \textbf{Understandability} through sparse concept bottlenecks, they may incur a higher cost due to the additional concept-processing layers.
These comparisons underscore that, to date, no single metric or method can satisfy all desiderata simultaneously; rather, a holistic evaluation across all desiderata is essential to provide a comprehensive profile of an XAI method's practical utility.

\subsection{Summary}
\label{sec:metric-summary}

This section systematizes the evaluation of visual XAI by bridging conceptual desiderata with practical metrics. 
The following key insights summarize the current landscape:

\begin{enumerate}[leftmargin=*]
\item \textbf{Core Desiderata:} Most metrics in visual XAI are proposed to fulfill four core desiderata: \textbf{Understandability}, \textbf{Fidelity}, \textbf{Continuity}, and \textbf{Efficiency}.
\item \textbf{Metric Priorities:} Localization metrics prioritize \textbf{Fidelity} through causal perturbations or spatial alignment, whereas semantic metrics emphasize \textbf{Understandability} and concept factuality.
\item \textbf{Toolkit Gap:} Existing toolkits provide mature pipelines for attribution-based methods but lack unified frameworks for concept-based evaluation.
\item \textbf{Evaluation Scope:} While metrics enable intra-group benchmarking, standardized protocols for comparing methods across dimensions remain underdeveloped.
\end{enumerate}

By summarizing these diverse desiderata and metrics, we provide a structured roadmap to guide researchers in selecting appropriate benchmarks for specific interpretability goals.
We anticipate the efforts of the research community to develop more objective, reliable, and universal metrics.

\section{XAI in Multimodal Models}
\label{sec:mllm}

Multimodal models \cite{wu2023multimodal,zhang2024mm} advance visual understanding by integrating textual and visual streams to tackle complex tasks.
While their cross-modal alignment facilitates user-friendly explanations, their massive scale and intricate fusion mechanisms pose substantial interpretability challenges.
This section discusses two perspectives: multimodal tools for interpretability (Sec.~\ref{sec:mllm-use}), leveraging these models to explain other systems; and interpretability of multimodal models (Sec.~\ref{sec:mllm-interp}), probing the models themselves.

\subsection{Multimodal Tools for Interpretability}
\label{sec:mllm-use}

In recent years, the rapid advancement of multimodal models has introduced novel technologies for research on the interpretability of visual recognition.
The alignment of visual and textual semantics within these models equips traditional visual recognition approaches with enhanced capabilities to explicitly represent semantics and provide natural language explanations.

For instance, conventional Concept Bottleneck Models~\cite{sawada2022concept, heidemann2023concept, lockhart2022learn} typically require predefined concept lists and manual annotations.
Leveraging multimodal alignment in models like CLIP~\cite{radford2021learning}, recent methods~\cite{oikarinen2023label, yang2023language, shang2024incremental, tan2024explain, fang2024cross} automate concept discovery and annotation, significantly reducing human labor.
Additionally, high-resolution and photorealistic image generation and editing models have enriched the data resources available for interpretability studies.
Recent works~\cite{farid2023latent, augustin2022diffusion, kim2023grounding, luo2023zero} employ these models to construct realistic probing and counterfactual datasets more efficiently than traditional synthesis.

Although large-scale models tend to be ``more black-box'', they present valuable opportunities for distilling embedded knowledge into interpretable frameworks.
Furthermore, powerful multimodal assistants enable cost-effective, objective evaluations, providing toolkits that reduce reliance on expensive user studies.

\subsection{Interpretability of Multimodal Models}
\label{sec:mllm-interp}

Interpreting current multimodal models is a challenging task due to their inherent complexity and the lack of established methods for probing their internal mechanisms.
Previous surveys~\cite{sun2024review,rodis2024multimodal,dang2024explainable,kazmierczak2025explainability} have summarized the related works.
According to existing consensus, research on the interpretability of multimodal models primarily focuses on \textbf{model interpretability} and \textbf{inference interpretability}.

\textbf{Model interpretability} pertains to the internal structure of these models.
Much of the existing work concentrates on vision transformers, such as through token analysis~\cite{kim2024vision,neo2024towards} and embedding analysis~\cite{verma2024cross,ramesh2022investigation}.
Traditional post-hoc interpretability methods, particularly attention-based techniques like GradCAM~\cite{selvaraju2017grad}, remain effective in most scenarios.
Notably, input-based probing methods—which are model-agnostic—play a crucial role in elucidating model behaviors~\cite{lindstrom2021probing,salin2022vision}.
Additional synthetic data is employed to mitigate data leakage and to better assess the decision-making processes of multimodal pre-trained models~\cite{fu2024blocks}.

It is worth noting that \textbf{inference interpretability} has emerged as an additional research focus for interpreting large-scale multimodal models.
Recent advances in Chain-of-Thought (CoT) techniques~\cite{gao2025interleaved,xu2024llava} have revitalized efforts to interpret the reasoning processes.
CoT refers to a methodology that prompts models to generate explicit intermediate reasoning steps in natural language, thereby enhancing performance on complex inference tasks.
Nevertheless, CoT methods are essentially generative in nature; while they facilitate user understanding, they still lack guarantees regarding correctness at the model level.

Overall, XAI for multimodal models remains in its early stages.
Significant efforts are needed to develop robust methods for interpreting these complex models to alleviate visual misperception, linguistic bias, or flawed cross-modal fusion.
Furthermore, as generative explanations like CoT become prevalent, establishing rigorous metrics to ensure their faithfulness to the internal model logic remains a critical challenge.
As multimodal models continue to evolve, developing robust, high-fidelity interpretive frameworks will be essential to ensuring transparency and trustworthiness in their real-world applications.

\section{Application and Discussion}
\label{sec:discussion}

Previously described XAI methods not only play an important role in revealing the mechanisms of visual recognition models but also are extensively utilized in various visual tasks and real-world applications.
In this section, we introduce several applications of XAI, highlighting its transformative impact across different fields.
Next, we will have a brief discussion on the trends, challenges and opportunities of XAI in visual recognition.

\subsection{XAI in Visual Tasks}
\label{sec:discussion-visual-task}

XAI has found applications across diverse visual tasks.
In data-scarce scenarios, XAI enhances model generalization and interpretability.
For instance, methods like LRP~\cite{bach2015pixel} and representation learning frameworks based on ProtoPNet~\cite{chen2019looks} provide explanations and jointly learn global and local features, improving few-shot classification performance~\cite{sun2021explanation, xu2020attribute}.
XAI also aids in interpreting and manipulating generative models, as frameworks like InterFaceGAN~\cite{shen2020interfacegan} and Network Dissection~\cite{bau2020understanding} enable editing of GAN-generated images and a deeper understanding of latent representations.
Advances in XAI, such as Relevance-based Neural Freezing~\cite{weber2023beyond} and the Reveal to Revise framework~\cite{pahde2023reveal}, improve model reliability by mitigating catastrophic forgetting and identifying spurious behaviors~\cite{ede2022explain}.
These examples illustrate XAI's potential to boost both model performance and user engagement across various visual tasks.

\subsection{XAI in Real-world Applications}
\label{sec:discussion-realworld}

\begin{table*}[t]
\caption{Summary of visual XAI surveys on real-world application.}
\label{tab:rs-c}
\begin{NiceTabular}{c>{\hspace{-2pt}}c<{\hspace{-2pt}}ccX}[width=\textwidth,cell-space-limits=0.5ex]
\CodeBefore
\rowcolor{gray!20}{1}
\rowlistcolors{2}{white,gray!10}[restart,cols={2-5}]
\Body
\toprule
\textbf{Group} & \textbf{Ref.} & \textbf{Year} & \textbf{Literature} & \textbf{Description} \\
\midrule
\Block[v-center]{3-1}{Medical Imaging} & \cite{klauschen2024toward} & 2024 & 2015-2023 & Explore diagnostic pathology: classification, biomarker quantification, transparency, solutions \\
& \cite{chaddad2023survey} & 2023 & 2019-2022 & Survey XAI techniques, categorize challenges, and suggest directions in medical imaging \\
& \cite{tjoa2020survey} & 2020 & 2015-2020 & Categorize AI interpretability approaches to guide cautious application in medical practices \\
\midrule
\Block[v-center]{2-1}{\shortstack{Industry /\\Manufacturing}} & \cite{li2023survey} & 2023 & 2016-2023 & Survey explainable anomaly detection: techniques, taxonomy, ethics, and guidance \\
& \cite{ahmed2022artificial} & 2022 & 2018-2021 & Survey XAI methods in Industry 4.0 for autonomous decision-making and transparency \\
\midrule
\Block[v-center]{2-1}{Smart City} & \cite{javed2023survey} & 2023 & 2018-2023 & Survey XAI in smart cities, focusing on use cases, challenges, and research directions \\
& \cite{kok2023explainable} & 2023 & 2018-2023 & Examine XAI in IoT: transparent models, challenges, and foresee future directions \\
\midrule
\Block[v-center]{2-1}{Cybersecurity} & \cite{charmet2022explainable} & 2022 & 2018-2022 & Survey XAI in cybersecurity: applications, security concerns, challenges, and future directions \\
& \cite{capuano2022explainable} & 2022 & 2018-2022 & Conduct study on XAI in cybersecurity: applications, challenges, methods, and the future \\
\bottomrule
\end{NiceTabular}
\end{table*}

The integration of XAI has transitioned from a theoretical preference to a fundamental necessity.
The following sections synthesize the current landscape of XAI applications as summarized in Tab.~\ref{tab:rs-c}.

\subsubsubsection{Medical Imaging and Healthcare}
In the realm of medical image analysis, explainability is no longer a luxury but a clinical requirement.
Advanced architectures, such as vision transformers and deep learning models, are being deployed to detect brain tumors, intracranial hemorrhage, and breast cancer.
To ensure these systems are trustworthy, researchers have focused on categorizing interpretability approaches to guide cautious application in medical practices \cite{tjoa2020survey}.
By integrating post-hoc explanation methods and example-based reasoning, current systems provide clinicians with interpretable evidence for diagnostic pathology and biomarker quantification \cite{klauschen2024toward, chaddad2023survey}.

\subsubsubsection{Industry 4.0 and Manufacturing}
In the industrial sector, XAI serves as a vital layer for safety and accountability.
Within Industry 4.0, XAI methods are utilized for autonomous decision-making and transparency \cite{ahmed2022artificial}.
A focus has been placed on explainable anomaly detection, where new taxonomies and ethical guidelines help researchers navigate the complexity of industrial sensor data while maintaining model performance \cite{li2023survey}.

\subsubsubsection{Smart Cities and IoT}
As urban environments become increasingly data-driven, XAI is redefining how AI interprets visual and sensor data in smart cities.
Research in this domain focuses on deciphering complex use cases and addressing the challenges of decentralized Internet of Things (IoT) networks \cite{javed2023survey}.
By fostering transparent models, these applications aim to align model transparency with societal perspectives to ensure public trust in smart infrastructure \cite{kok2023explainable}.

\subsubsubsection{Cybersecurity and Safety-Critical Systems}
In specialized tasks such as deepfake face detection and cybersecurity monitoring, models must not only provide a result but also characterize the features driving the decision.
XAI is being applied to address security concerns and future directions in threat detection \cite{charmet2022explainable, capuano2022explainable}.
Similarly, in autonomous driving and remote sensing, interpretable models allow researchers to translate complex data into human-understandable logic, ensuring that AI systems remain safe, ethical, and effective.

\subsection{Trends, Challenges, and Opportunities}
\label{sec:discussion-trend}

Despite the significant advancements in XAI for visual recognition, the practical application of these methods remains relatively limited compared to black-box models.

We begin by outlining the emerging trends in this field, framed through the lens of our proposed taxonomy.
Regarding \textbf{intent}, while post-hoc (passive) methods remain prevalent, self-interpretable (active) models are increasingly viewed as providing a more intrinsic solution to the black-box dilemma~\cite{rudin2022interpretable}.
In the \textbf{object} dimension, local, semi-local, and global explanations address different application scenarios, and some recent efforts have aimed to unify these approaches within a single framework~\cite{schrouff2021best,visani2024gleams}.
\textbf{Presentation} constitutes the most distinctive dimension separating visual XAI from generic XAI, as it provides the most direct and intuitive experience for human users.
The forms of explanation vary according to whether the objective is \textbf{localization interpretability} or \textbf{semantic interpretability}; some forms are applicable to both, and recent studies have focused on achieving both types to ensure comprehensive explanations~\cite{dani2023devil,wan2024interpretable}.
Finally, \textbf{methodology} is closely intertwined with the aforementioned dimensions, and the correspondence with the levels of the ladder of causation highlights the diverse perspectives XAI methods adopt when investigating model behaviors.

Despite these advancements, the application scope of interpretable visual models remains significantly narrower than that of state-of-the-art black-box models.
This gap stems from the relative lag in interpretability research compared to pure recognition performance, alongside concerns regarding the potential accuracy trade-offs inherent in certain interpretable architectures.
Some of these challenges, such as performance limitations, may be alleviated by further technological advancements~\cite{rudin2022interpretable}.
Furthermore, the wider adoption of interpretable models depends on users' increasing demand for model reliability.
We believe that as AI models become integrated into more aspects of daily life, the importance of interpretability will continue to grow.

\section{Conclusion}
\label{sec:conclusion}

Despite the rapid evolution and widespread adoption of visual recognition, trust remains a pivotal concern in high-stakes domains, underscoring the necessity for model interpretability.
This survey systematically reviews the landscape of interpretable visual recognition, categorizing existing methods across four core dimensions: \textbf{intent}, \textbf{object}, \textbf{presentation}, and \textbf{methodology}.
The proposed taxonomy serves as a framework for researchers and practitioners to align user requirements with appropriate interpretability techniques. 
Furthermore, we analyze the fundamental desiderata of XAI, consolidate existing evaluation metrics, and explore the interpretability of emerging multimodal models alongside their practical applications. 
We anticipate that this overview will facilitate the deployment of transparent AI and enhance human trust in visual recognition systems.

\section*{Acknowledgments}

This work is partially supported by National Key R\&D Program of China No. 2021ZD0111901, and Natural Science Foundation of China under contracts Nos. U21B2025.

\bibliographystyle{IEEEtran}
\bibliography{main}
 
\vfill
\end{document}